\documentclass{article}
\usepackage{amsmath,epsfig,bm, amssymb,subfigure,graphicx,algorithm,algorithmic,color,natbib}
\newtheorem{defi}{Definition}

\newtheorem{theo}[defi]{Theorem}
\newtheorem{coro}[defi]{Corollary}

\newcommand{\proofend}{\hfill$\Box$\vspace{2mm}}

\newcommand{\mathbbE}{\mathbb{E}}

\newcommand{\mathbbR}{\mathbb{R}}

\newcommand{\boldzero}{{\boldsymbol{0}}}
\newcommand{\boldone}{{\boldsymbol{1}}}

\newcommand{\boldA}{{\boldsymbol{A}}}

\newcommand{\boldI}{{\boldsymbol{I}}}

\newcommand{\boldX}{{\boldsymbol{X}}}
\newcommand{\boldY}{{\boldsymbol{Y}}}

\newcommand{\bolda}{{\boldsymbol{a}}}
\newcommand{\boldb}{{\boldsymbol{b}}}

\newcommand{\boldu}{{\boldsymbol{u}}}

\newcommand{\boldx}{{\boldsymbol{x}}}
\newcommand{\boldy}{{\boldsymbol{y}}}
\newcommand{\boldz}{{\boldsymbol{z}}}

\newcommand{\boldalpha}{{\boldsymbol{\alpha}}}

\newcommand{\boldeta}{{\boldsymbol{\eta}}}

\newcommand{\boldmu}{{\boldsymbol{\mu}}}

\newcommand{\boldSigma}{{\boldsymbol{\Sigma}}}

\newcommand{\calD}{{\mathcal{D}}}

\newcommand{\calF}{{\mathcal{F}}}

\newcommand{\calN}{{\mathcal{N}}}

\newcommand{\calS}{{\mathcal{S}}}

\usepackage{url,multirow}

\advance\oddsidemargin-.75in
\date{\today}
\title{Post Selection Inference with \\Incomplete Maximum Mean Discrepancy Estimator} 

\author{
Makoto Yamada$^{1,2,5}\thanks{Equal contribution}$, Denny Wu$^{3}\footnotemark[1]$, , Yao-Hung Hubert Tsai$^3$\\
Ichiro Takeuchi$^{1,4}$, Ruslan Salakhutdinov$^3$, Kenji Fukumizu$^{1,5}$ \\
$^1$RIKEN AIP, Japan \\
$^2$JST PRESTO, Japan \\
$^3$Carnegie Mellon University, USA \\
$^4$Nagoya Institute of Technology, Japan\\
$^5$Institute of Statistical Mathematics, Japan \\
}

\begin{document}
\maketitle

\begin{abstract}
Measuring divergence between two distributions is essential in machine learning and statistics and has various applications including binary classification, change point detection, and two-sample test. Furthermore, in the era of big data, designing divergence measure that is interpretable and can handle high-dimensional and complex data becomes extremely important. In the paper, we propose a post selection inference (PSI) framework for divergence measure, which can select a set of statistically significant features that discriminate two distributions. Specifically, we employ an additive variant of maximum mean discrepancy (MMD) for features and introduce a general hypothesis test for PSI. A novel MMD estimator using the incomplete U-statistics, which has an asymptotically normal distribution (under mild assumptions) and gives high detection power in PSI, is also proposed and analyzed theoretically. Through synthetic and real-world feature selection experiments, we show that the proposed framework can successfully detect statistically significant features. Last, we propose a sample selection framework for analyzing different members in the Generative Adversarial Networks (GANs) family. 
\end{abstract}

\section{Introduction} 
Computing the divergence between two probability distributions is fundamental to machine learning and has many important applications such as binary classification \citep{friedman2001elements}, change point detection \citep{yamada2013change,liu2013change}, two-sample test \citep{gretton2012kernel,yamada2013relative}, and generative models such as generative adversarial networks (GANs) \citep{goodfellow2014generative,Li2015mmd,nowozin2016f}, to name a few. Recently, interpreting the difference between distributions has become an important task in applied machine learning \citep{mueller2015principal,jitkrittum2016interpretable} since it can facilitate scientific discovery. For instance, in biomedical binary classification tasks, it is common to analyze which variables or features are different between two different distributions (classes). 

\begin{figure*}[t!]
\begin{center}
\begin{minipage}[t]{0.9\linewidth}
\centering
{\includegraphics[width=0.99\textwidth]{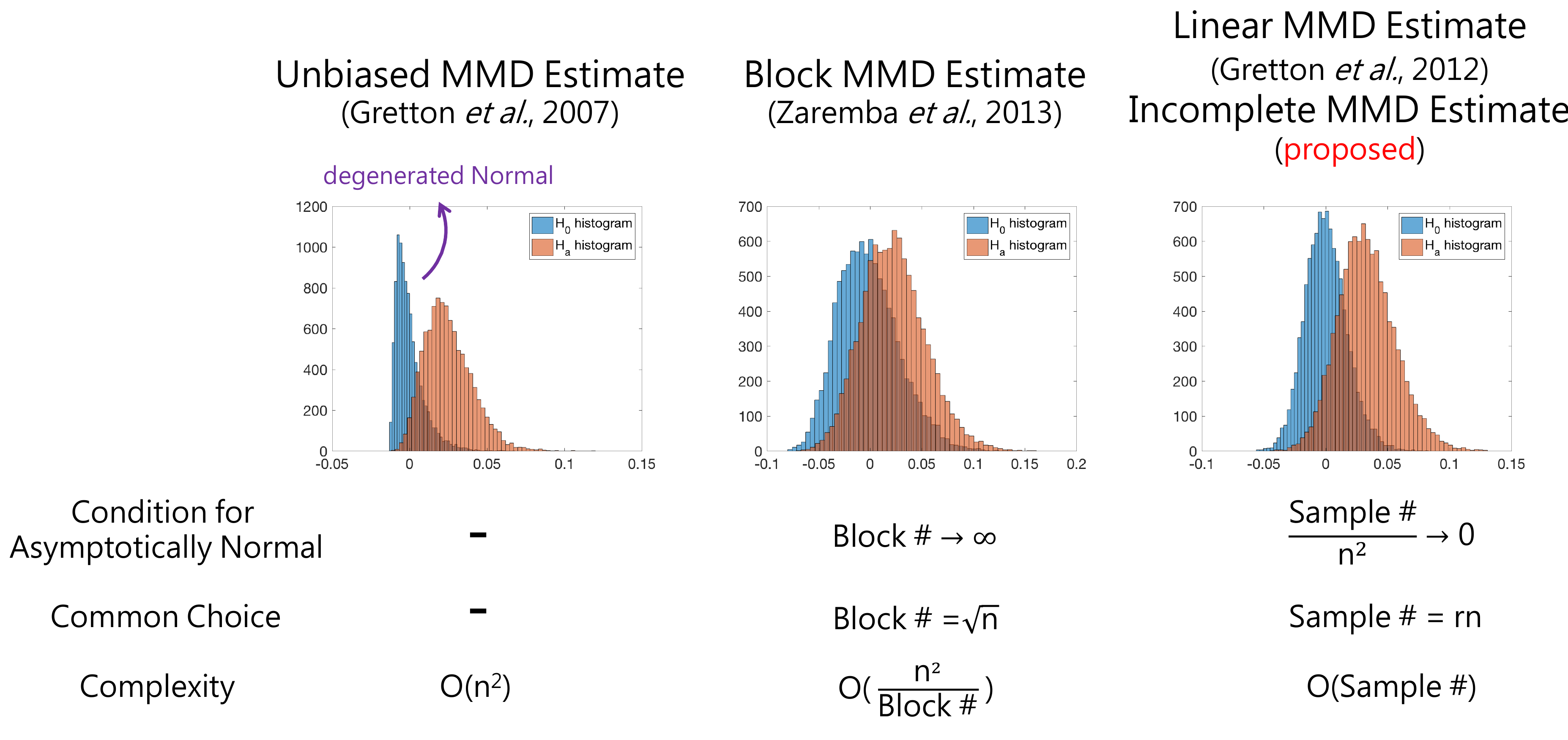}} \\ \vspace{-0.10cm}
\end{minipage}
 \caption{Comparison between Unbiased, Block, Linear, and Incomplete MMD estimation.}
    \label{fig:illustration}
\end{center}
\vspace{-.1in}
\end{figure*}

The simplest approach to measure divergence between two probability densities would be parametric methods. For example, $t$-test can be used if the distributions to be compared are known and well-defined \citep{anderson2001new}. However, in many real-world problems, the property of distribution is not known \emph{a priori}, and therefore model assumptions are likely to be violated. In contrast, non-parametric methods can be applied to any distributions without prior assumptions.  The maximum mean discrepancy (MMD) \citep{gretton2012kernel} is an example of non-parametric discrepancy measures and is defined as the difference between the mean embeddings of two distributions in a reproducing kernel Hilbert space (RKHS). Due to the mean embeddings in RKHS, all moment information is stored. Thus in comparing the means of a characteristic kernel, MMD captures non-linearity in data and can be computed in a closed-form. However, since MMD considers the entire $d$ dimensional vector, it is hard to interpret how individual features contribute to the discrepancy. 
  
To deal with the interpretability issue, divergence measure with feature selection has been actively studied \citep{yamada2013change,mueller2015principal,jitkrittum2016interpretable}. For instance, in \citet{mueller2015principal}, the Wasserstein divergence is employed as a divergence measure and $\ell_1$ regularizer is used for feature selection. However, these approaches focus on detecting a set of features that discriminate two distributions. But for scientific discovery applications such as biomarker discovery, it might be preferable to test the significance of each selected feature (e.g., one biomarker). A naive approach would be to select features from one dataset and then test the selected features using the same data. However, in such case, \emph{selection bias} is included and thus the false positive rate cannot be controlled. Therefore, it is crucial in hypothesis testing that the selection event should be taken into account to correct the bias. To the best of our knowledge, there is not yet an existing framework that tests the significance of selected features that distinguish between two different distributions on the same dataset.
  
  In this paper, we propose \texttt{mmdInf}, a \emph{post selection inference} (PSI) algorithm for distribution comparison, which finds a set of \emph{statistically significant} features that discriminate between two distributions. \texttt{mmdInf} enjoys several compelling properties. First, it is a non-parametric method based on kernels, and thus it can detect features that distinguish various types of distributions. Second, the proposed framework is general and can be applied to not only feature selection but also other distribution comparison problems, such as change point detection and dataset selection. In \texttt{mmdInf}, we employ the recently developed PSI algorithm \citep{lee2016exact} and use MMD as a divergence measure. However, the standard empirical squared MMD estimator has a degenerated null distribution, which violates the requirement of a proper Normal distribution for the current PSI algorithm. To address this issue, we apply the normally distributed block estimate \citep{zaremba2013b} and the linear estimate \citep{gretton2012kernel} of MMD. Furthermore, we propose a new empirical estimate of MMD based on the incomplete U-statistics (incomplete MMD estimator) \citep{blom1976some,janson1984asymptotic,lee1990u} and show that it asymptotically follows the normal distribution, and has much greater detection power (compared to the existing estimators of MMD) in PSI. Finally, we propose a framework to analyze different members in the GAN \citep{goodfellow2014generative} family based on \texttt{mmdInf}. We elucidate the theoretical properties of the incomplete U-statistics estimate of MMD and show that \texttt{mmdInf} can successfully detect significant features through feature selection experiments.
  
The contributions of our paper are summarized as follows:
\begin{itemize}
\item We propose a non-parametric PSI algorithm \texttt{mmdInf} for distribution comparison. 
\item We propose the incomplete MMD estimator and investigate its theoretical properties (see Figure~\ref{fig:illustration}).
\item {We propose a sample selection framework based on \texttt{mmdInf} that can be used for analyzing generative models}.
\end{itemize}

\section{Related Work}
There exist a number of divergence measures including the Kullback-Leibler divergence \citep{cover2012elements}, $f$-divergence \citep{ali1966general}, $\alpha$-divergence \citep{renyi1961measures}, K-nearest neighbor approaches \citep{poczos2011estimation}, density-ratio based approaches \citep{yamada2013relative,sugiyama2008direct,kanamori2009least}, etc. Among these divergence measures, we are interested in Kernel Mean Embedding (KME) approaches. For instance, the maximum mean discrepancy (MMD) \citep{gretton2012kernel} is a widely used kernel based divergence that compares the means of two distributions in a reproducing kernel Hilbert space (RKHS). Although these divergence measures can be used for testing the discrepancy between $p(\boldx)$ and $q(\boldx)$, it is hard to test the significance of one of $k$ selected features from the entire $d$ features due, where the setup is useful for scientific discovery tasks such as biomarker discovery. Recently, \citep{li2015m} proposed a MMD-based change point detection algorithm, which can compute the $p$-value of the maximum MMD score. However, this method can only test the maximum MMD score (i.e., one-feature). Thus, it is not clear whether the approach can be extended to feature selection in general. 

A novel testing framework for the post selection inference (PSI) has been recently proposed \citep{lee2016exact}, in which statistical inference after feature selection with Lasso \citep{tibshirani1996regression} is investigated. This work shows that statistical inference conditioned on the selection event can be done for linear regression models with Gaussian noise if the  selection event can be written as a set of linear constraints. This result is known as the Polyhedral Lemma. However, the PSI algorithm needs to assume a Gaussian response, which is a relatively strong assumption, and consequently it cannot be directly applied for non-Gaussian output problems such as classification. To deal with this issue, a kernel based PSI algorithm (\texttt{hsicInf}) for \emph{independence test} has been proposed \citep{yamada2018post}, in which the Hilbert-Schmidt Independence Criterion (HSIC) \citep{gretton2005measuring} is employed to measure the independence between input and its output, and thus significance test can be performed on non-Gaussian data. In this paper, we propose an alternative kernel based inference algorithm for \emph{distribution comparison} called \texttt{mmdInf}, which can be used for both feature selection and binary classification.  

In Sections 3.4 and 5.5, we manifest how we analyze different members in the GANs \citep{goodfellow2014generative} family. Here, we discuss several evaluation metrics that have been proposed to compare generative models. For example, the inception scores \citep{salimans2016improved} and the mode scores \citep{che2016mode} measure the quality and diversity of the generated samples, but they were not able to detect overfitting and mode dropping / collapsing for generated samples. The Frechet inception distance (FID) \citep{heusel2017gans} defines a score using the first two moments of the real and generated distributions, whereas the classifier two-sample tests \citep{lopez2016revisiting} considers the classification accuracy of a binary classifier as a statistic for two-sample test. Although the above metrics are reasonable in terms of discriminability, robustness, and efficiency, the distances between samples are required to be computed in a suitable feature space. We can also use the kernel density estimation (KDE); or more recently, \cite{wu2016quantitative} proposed to apply the annealed importance sampling (AIS) to estimate the likelihood of the decoder-based generative models. Nevertheless, these approaches need the access to the generative model for computing the likelihood, which are less favorable comparing to the model agnostic approaches which rely only on a finite generated sample set. On the other hand, the maximum mean discrepancy (MMD) \citep{gretton2012kernel} is preferred against its competitors \citep{sutherland2016generative,huang2018an}. Therefore, we propose the \texttt{mmdInf} based GANs analysis framework.  

\section{Proposed Method (\texttt{mmdInf})}
In this section, we introduce a PSI algorithm with the maximum mean discrepancy (MMD) \citep{gretton2012kernel}.  We first formulate the PSI problem and then propose the MMD inference algorithm \texttt{mmdInf}.
 
\subsection{Problem Formation}
Suppose we are given independent and identically distributed (i.i.d.) samples $\boldX = \{\boldx_i \}_{i = 1}^{m} \in \mathbbR^{d \times m}$ from a $d$-dimensional distribution $p$  and i.i.d. samples $\boldY = \{\boldy_j \}_{j = 1}^{n} \in \mathbbR^{d \times n}$ from another $d$-dimensional distribution $q$. Our goal is to find $k < d$ features that differentiate between $\boldX$ and $\boldY$ and test whether each selected feature is {\em statistically significant}.

Let $\calS$ be a set of selected features, we consider the following hypothesis test:
\begin{itemize}
\item $H_{0}$: $\sum_{s = 1}^d \eta_s \widehat{D}(\boldX^{(s)},\boldY^{(s)})  = 0 ~|~   \text{$\calS$ was selected}$, 
\item $H_{1}$: $\sum_{s = 1}^d \eta_s \widehat{D}(\boldX^{(s)},\boldY^{(s)}) ~ \neq 0 ~|~  \text{$\calS$ was selected}$,
\end{itemize}
where $\widehat{D}(\boldX^{(s)},\boldY^{(s)})$ is the estimated discrepancy measure for the selected feature $s$ and $\boldeta = [\eta_1, \ldots, \eta_d]^\top \in \mathbbR^{d}$ is an arbitrary pre-defined parameter.

To test the $s$-th feature, the hypothesis test for the selected feature $s$ can be written as
\begin{itemize}
\item $H_{0}$: $\widehat{D}(\boldX^{(s)},\boldY^{(s)})  = 0 ~|~   \text{$\calS$ was selected}$, 
\item $H_{1}$: $\widehat{D}(\boldX^{(s)},\boldY^{(s)}) ~ \neq 0 ~|~  \text{$\calS$ was selected}$,
\end{itemize}
where 
\begin{align*}
\boldeta &= [{0~\cdots 0~\underbrace{1}_{s}~0~\cdots 0}]^\top \in \mathbbR^{d}.
\end{align*}

\subsection{Post Selection Inference (PSI)}
We employ the post-selection inference (PSI) framework \citep{lee2016exact} to test the hypothesis.

\begin{theo} \label{theo-psi}\citep{lee2016exact}
Suppose that $\boldz \sim \calN(\boldmu,\boldSigma)$, and the feature selection event can be expressed as $\boldA \boldz \le \boldb$ for some matrix $\boldA$ and vector $\boldb$, then for any given feature represented by $\boldeta \in \mathbbR^n$ we have
\begin{equation*}
F_{\boldeta^\top\boldmu,\boldeta^\top\boldSigma\boldmu}^{[V^{-}(\boldA,\boldb),V^{+}(\boldA,\boldb)]}(\boldeta^\top \boldz) \quad | \quad
\boldA\boldz \le \boldb \sim \textnormal{Unif}(0,1),
\end{equation*}
where $F_{\mu,\sigma^2}^{[a,b]}(x)$ is the cumulative distribution function (CDF) of a truncated normal distribution truncated at [a,b], and $\Phi$ is the CDF of standard normal distribution with mean $\mu$ and variance $\sigma^2$. Given that $\boldalpha = \boldA\frac{\boldSigma\boldeta}{\boldeta^\top\boldSigma\boldeta}$, the lower and upper truncation points are given by
\begin{align*}
V^{-}(\boldA,\boldb) &= \max_{j: \boldalpha_{j} < 0} \frac{b_{j} - (\boldA \boldz)_{j}}{\boldalpha_{j}} + \boldeta^\top \boldz, \\
V^{+}(\boldA,\boldb) &= \min_{j: \boldalpha_{j} > 0} \frac{\boldb_{j} - (\boldA\boldz)_{j}}{\boldalpha_{j}} + \boldeta^\top \boldz.
\end{align*}
\end{theo}

\paragraph{Marginal Screening with Discrepancy Measure:} 
Assume we have an estimate of a discrepancy measure for each feature: $\boldz = [\widehat{D}(\boldX^{(1)}, \boldY^{(1)}), \ldots, \widehat{D}(\boldX^{(d)}, \boldY^{(d)})]^\top \in \mathbbR^{d} \sim \calN(\boldmu, \boldSigma)$, where $\widehat{D}(\cdot, \cdot)$ {has large positive value when two distribution are different}. In this paper, we select the top-$k$ features with largest discrepancy scores. We denote the index set of the selected $k$ features by $\calS$, and that of the unselected $\bar{k} = d -k$ features by $\bar{\calS}$. This feature selection event can be characterized by
\begin{align*}
\widehat{D}(\boldX^{(s)}, \boldY^{(s)}) \geq \widehat{D}(\boldX^{(\ell)}, \boldY^{(\ell)}), ~~\textnormal{for all}~ (s, \ell) \in \calS \times \bar{\calS}.
\end{align*}
Note that we have in total $k \cdot \bar{k}$ constraints.

The selection event can be rewritten as
\begin{align*}
\boldA_{s,\ell} \boldz \leq 0, ~~\textnormal{for all}~ (s, \ell) \in \calS \times \bar{\calS},
\end{align*}
where
\begin{align*}
\boldA_{s,\ell} &= [{0~\cdots 0~\underbrace{-1}_{s}~0~\cdots 0~\underbrace{1}_{\ell}~0 \cdots 0}] 
\end{align*}
 and $\boldA_{s,\ell}^\top \in \mathbbR^d$ is a row vector of $\boldA \in \mathbbR^{(k \cdot \bar{k}) \times d}$. Under such construction, $\boldA \boldz \le \boldb$ can be satisfied by setting $\boldb = \boldzero$.

\subsection{Maximum Mean Discrepancy (MMD)}
We employ the maximum mean discrepancy (MMD) \citep{gretton2012kernel} as divergence measure. 

Let $\calF$ be the unit ball in a reproducing kernel Hilbert space (RKHS) and $k(\boldx,\boldx')$ the corresponding positive definite kernel, the squared population MMD is defined as
\begin{align*}
&\text{MMD}^{2}[\calF,p,q] = \textbf{E}_{\boldx,\boldx'}[k(\boldx,\boldx')] - 2\textbf{E}_{\boldx,\boldy}[k(\boldx,\boldy)] + \textbf{E}_{\boldy,\boldy'}[k(\boldy,\boldy')],
\end{align*}
where $\textbf{E}_{\boldx,\boldy}$ denotes the expectation over independent random variables $\boldx$ with distribution $p$ and $\boldy$ with distribution $q$. It has been shown that if the kernel function $k(\boldx,\boldx')$ is characteristic, then $\text{MMD}[\calF,\boldX,\boldY]= 0$ if and only if $p = q$ \citep{gretton2012kernel}.

In the following, we introduce two existing MMD estimators and then propose our new MMD estimator based on the incomplete U-statistics.

\vspace{.05in}
\noindent {\bf (Complete) U-statistics estimator \citep{gretton2012kernel}:}
We use the Gaussian kernel: 
\[
k(\boldx,\boldx') = \exp\left(-\frac{\|\boldx - \boldx'\|_2^2}{2\sigma_x^2}\right),
\]
where  $\sigma_x > 0$ is the Gaussian width. 

Then, the \emph{complete} U-statistics of MMD is defined as
\begin{align*}
&\textnormal{MMD}_u^2[\calF, \boldX, \boldY] =　\frac{1}{m(m-1)}\!\sum_{i=1}^m\!\sum_{j \neq i}^m k(\boldx_i,\boldx_j) \!+\! \frac{1}{n(n-1)}\!\sum_{i=1}^n\!\sum_{j \neq i}^n k(\boldy_i,\boldy_j) - \frac{1}{mn} \sum_{i = 1}^m \sum_{j = 1}^n k(\boldx_i,\boldy_j).
\end{align*}

In particular, when $m = n$, we can write the estimate as
\begin{align*}
\textnormal{MMD}_u^2[\calF, \boldX, \boldY] =\frac{1}{m(m-1)} \sum_{i \neq j}h(\boldu_i, \boldu_{j}),
\end{align*}
where
\begin{align*}
h(\boldu,\boldu') \!=\!k(\boldx,\boldx')\!+\! k(\boldy,\boldy')\!-\! k(\boldx,\boldy')\!-\! k(\boldx',\boldy)
\end{align*}
is the U-statistics kernel for MMD and $\boldu = [\boldx^\top~\boldy^\top]^\top \in \mathbbR^{2d}$. However, since the complete U-statistics estimator of MMD is degenerated under $p = q$ and does not follow normal distribution, this estimator cannot be used in the current PSI framework.

\vspace{.05in}
\noindent {\bf Block estimator \citep{zaremba2013b}:}
Let us partition $\boldX$ and $\boldY$ into $\frac{m}{B_1} = \frac{n}{B_2}$ blocks where each block consists of $B_1$ and $B_2$ samples:
\begin{align*}
\boldX &= [\boldX_1, \ldots, \boldX_{m/B_1}],~~\boldY = [\boldY_1, \ldots, \boldY_{n/B_2}].
\end{align*}
Here, we assume that the number of blocks $m/B_1 = n/B_2$ is an integer. Then, the block estimate of MMD is given by
\begin{equation*}
\textnormal{MMD}_{b}^{2}[\calF,\boldX,\boldY]=\frac{B_1}{n} \sum_{i=1}^{n/B_1}\textnormal{MMD}_{u}^{2}[\calF,\boldX_{i},\boldY_{i}],
\end{equation*}
where $\boldX_i$ and $\boldY_{i}$ are data of the $i$-th block. This estimator asymptotically follows the normal distribution when $B_1$ and $B_2$ are finite and $m$ and $n$ go to infinity. The block estimator can be used for PSI, but the variance and normality depends on the partition of $\boldX$ and $\boldY$. Specifically, when the total number of samples is small, then a small block size would result in high variance, whereas larger block size tends to result in non-Gaussian response.

\vspace{.05in}
\noindent {\bf Incomplete U-statistics estimator:}
The described problems of the block-estimator motivated us to design a new MMD estimator that is normally distributed and has smaller variance. We therefore propose an MMD estimator based on the incomplete U-statistics \citep{blom1976some,janson1984asymptotic,lee1990u}.  

 The incomplete U-statistics estimator of MMD is given by
\begin{align*}
&\textnormal{MMD}_{inc}^{2}[\calF,\boldX,\boldY]= \frac{1}{\ell}\sum_{(i,j) \in \calD} h(\boldu_i,\boldu_{j}),   
\end{align*}
where $\calD$ is a subset of $\calS_{n,k} = \{(i,j)\}_{i \neq j}$. $\calD$ can be fixed design or random design. In particular, if we design $\calD$ as
\[
\calD = \{(1,2),(3,4), \ldots, (n-1,n)\}
\]
and assume that $n$ is an even number, then the incomplete U-statistic corresponds to the linear-time MMD estimator \citep{gretton2012kernel}:
\begin{align*}
\text{MMD}_{lin}^2[\calF, \boldX,\boldY]　= \frac{2}{n} \sum_{i = 1}^{n/2} h(\boldu_i, \boldu_{i+1}).
\end{align*}

\subsection{Additional Applications (GANs analysis)}
The proposed PSI framework is general and can be used for not only feature selection but also sample selection. In generative modeling, the generated distribution should match the data distribution; in other words, the discrepancy between the generated data and real data should be small. We can therefore apply the selective inference algorithm and use the significance value to evaluate the generation quality. In this paper we apply \texttt{mmdInf} to compare the performance of GANs. We first select the model whose generated samples has the smallest MMD score with the real data and then perform the hypothesis test.

Let $\boldx^{(s)}_i \in \mathbbR^{p}$ be a feature vector generated by $s$-th GAN model with random seed $i$　 and $\boldy_j \in \mathbbR^{p}$ is a feature vector of an original image. Image features can be extracted by pre-trained Resnet \citep{he2016deep} or auto-encoders. The hypothesis test can be written as
\begin{itemize}
\item $H_{0}$: $\textnormal{MMD}_{inc}^2[\calF,\boldX^{(s)},\boldY] \! =\! 0 ~|~ \text{$s$ generates samples closest to the real distribution}$, 
\item $H_{1}$: $\textnormal{MMD}_{inc}^2[\calF,\boldX^{(s)},\boldY] \!\neq 0\! ~|~ \text{$s$ generates samples closest to the real distribution}$.
\end{itemize}
Since we want to test the best generator that minimizes the discrepancy between generated and true samples (e.g., low MMD score),  this sample selection event can be characterized by
\begin{align*}
\textnormal{MMD}_{inc}^2[\calF,\boldX^{(s)}, \boldY] \leq \textnormal{MMD}_{inc}^2[\calF,\boldX^{(\ell)}, \boldY], 
\end{align*}
where $ (s, \ell) \in \calS \times \bar{\calS}$.
 
\section{Theoretical Analysis of Incomplete MMD}
We investigate the theoretical properties of the incomplete MMD estimator under the random design with replacement.  For simplicity, we denote $\text{MMD}_{inc}^2[\calF,\boldX, \boldY] = \text{MMD}_{inc}^2$ and $\text{MMD}^2[\calF,p,q] = \text{MMD}^2$, respectively.

\begin{theo} 
\label{theo_mmd_normal1}
Let $n$ and $\ell$ tend to infinity such that $\gamma = \lim_{n,\ell \to \infty} n^{-(c+1)}\ell, ~ 0 \leq \gamma \leq \infty$. For sampling with replacement, we have

$\begin{cases}
      \ell^{\frac{1}{2}}(\textnormal{MMD}_{inc}^2 - \textnormal{MMD}^2) \xrightarrow[]{d} \calN(0, \sigma^2),  & \text{if}\ \gamma = 0. \\
      n^{\frac{c+1}{2}}(\textnormal{MMD}_{inc}^2 - \textnormal{MMD}^2) \xrightarrow[]{d} V, & \text{if}\ \gamma = \infty. \\
        \ell^{\frac{1}{2}}(\textnormal{MMD}_{inc}^2 \!-\! \textnormal{MMD}^2)  \!\xrightarrow[]{d}\! \gamma^{\frac{1}{2}}V + T, & \text{if}\ 0<\gamma<\infty. \\
\end{cases}$

where $V$ is the random variable of the limit distribution of $ n^{\frac{c+1}{2}}(\textnormal{MMD}_{u}^2 - \textnormal{MMD}^2)$,  $T$ is the random variable of $\calN(0, \sigma^2)$, $\sigma^2 = Var(h(\boldu,\boldu'))$,  and $T$ and $V$ are independent. %

\vspace{.05in}
\noindent {\bf Proof:} Use Corollary 1 of \citet{janson1984asymptotic} (or Theorem 1 of \citet{lee1990u}, pp. 200). {In MMD, $c = 1$ for $p=q$ and $c = 0$ for $p \neq q$.} \proofend
\end{theo}

\begin{coro} 
Assume $\lim_{n,\ell \to \infty} n^{-2}\ell = 0$ and $0 <\gamma = \lim_{n,\ell \to \infty} n^{-1}\ell< \infty$. For sampling with  replacement, the incomplete U-statistics estimator of MMD is asymptotically normally distributed as

$\begin{cases}
      \ell^{\frac{1}{2}}\textnormal{MMD}_{inc}^2  \xrightarrow[]{d} \calN(0, \sigma^2),  & \text{if}\ p=q. \\
       \ell^{\frac{1}{2}} (\textnormal{MMD}_{inc}^2 - \textnormal{MMD}^2)   \xrightarrow[]{d}  \calN(0, \sigma^2 + \gamma\sigma_u^2), &  \text{if}\ p \neq q. 
\end{cases}$

where  $\sigma^2 = Var(h(\boldu,\boldu'))$ and $\sigma_u^2 = 4(\mathbbE_{\boldu}[(\mathbbE_{\boldu'}[h(\boldu, \boldu')]] - \mathbbE_{\boldu,\boldu'}[h(\boldu,\boldu')])^2)$.

\vspace{.05in}
\noindent {\bf Proof:} Under $p = q$ ($c = 1$), since $\lim_{n,\ell \to \infty} n^{-2}\ell = 0$ and $\textnormal{MMD}^2 = 0$, we can immediately obtain the limit distribution by Theorem~\ref{theo_mmd_normal1}. Under $p \neq q$ ($c = 0$), $\textnormal{MMD}_u$ converges in distribution to a Gaussian according to \citep{gretton2007kernel}
\begin{align*}
n^{\frac{1}{2}}(\textnormal{MMD}^2_u - \textnormal{MMD}^2) \xrightarrow[]{d} \calN(0, \sigma_u^2)
\end{align*}
where  $\sigma_u^2 = 4(\mathbbE_{\boldu}[(\mathbbE_{\boldu'}[h(\boldu, \boldu')]] - \mathbbE_{\boldu,\boldu'}[h(\boldu,\boldu')])^2)$. Based on Theorem \ref{theo_mmd_normal1}, under the given assumption, we can obtain the distribution of MMD$^2_{inc}$ since $T$ and $V$ are independent.  \proofend
\end{coro}

\begin{coro} 
Assume $\lim_{n,\ell \to \infty} n^{-1}\ell = 0$. For sampling with replacement, the incomplete U-statistics estimator of MMD is asymptotically normally distributed as
\begin{align*}
      \ell^{\frac{1}{2}}(\textnormal{MMD}_{inc}^2  - \textnormal{MMD}^2) \xrightarrow[]{d} \calN(0, \sigma^2).
\end{align*}
\vspace{.05in}
\noindent {\bf Proof:} Since $\lim_{n,\ell \to \infty} n^{-1}\ell = 0$ and $\lim_{n,\ell \to \infty} n^{-2}\ell = 0$, the limit distribution of $\ell^{1/2}(\textnormal{MMD}_{inc}^2  - \textnormal{MMD}^2)$ is $\calN(0, \sigma^2)$ based on Theorem \ref{theo_mmd_normal1}. \proofend
\end{coro}

Thus, in practice, by setting $\ell \ll n^2$, the incomplete estimator is asymptotically normal and therefore can be applied in PSI. More specifically, we can set $\ell = r n \ll n^2$, where $r$ is a small constant. In practice, we found that $r = 10$ works well in general.

\begin{figure*}[t!]
\begin{center}
\begin{minipage}[t]{0.245\linewidth}
\centering
{\includegraphics[width=0.99\textwidth]{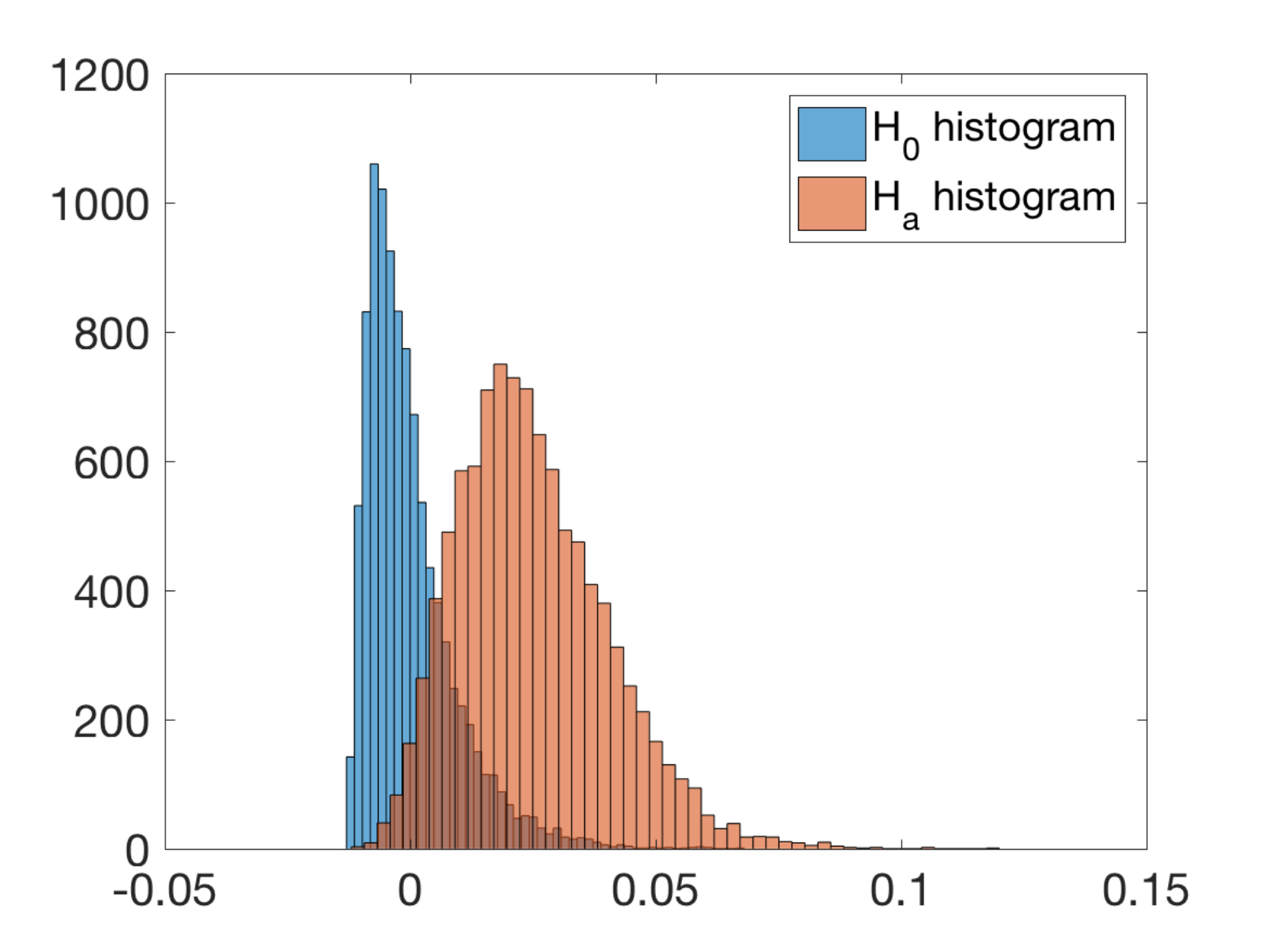}} \\ \vspace{-0.10cm}
(a) MMD$_{u}$.
\end{minipage}
\begin{minipage}[t]{0.245\linewidth}
\centering
 {\includegraphics[width=0.99\textwidth]{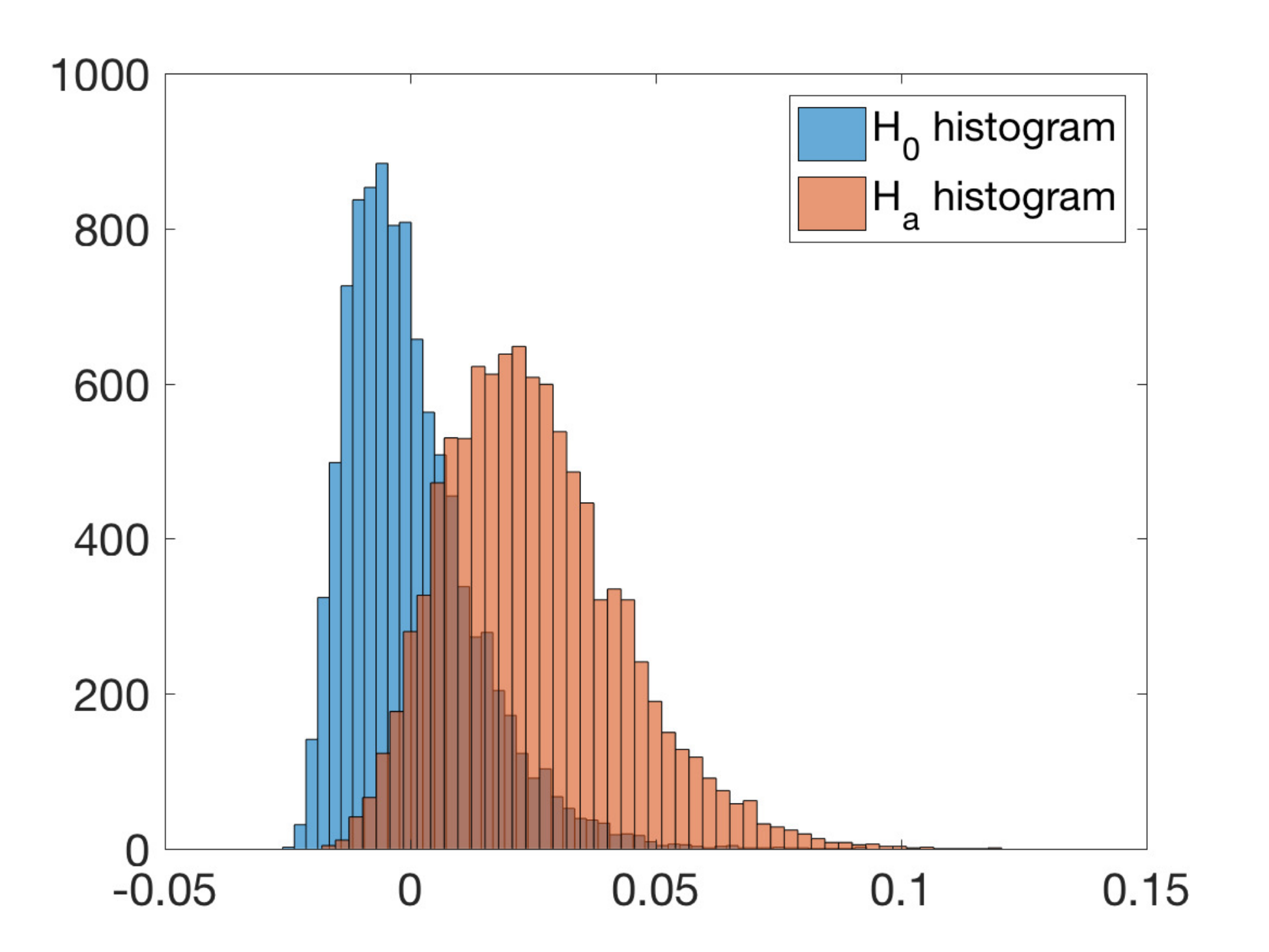}} \\ \vspace{-0.10cm}
(b) MMD$_{b},  B = 100$.
\end{minipage}
\begin{minipage}[t]{0.245\linewidth}
\centering
{\includegraphics[width=0.99\textwidth]{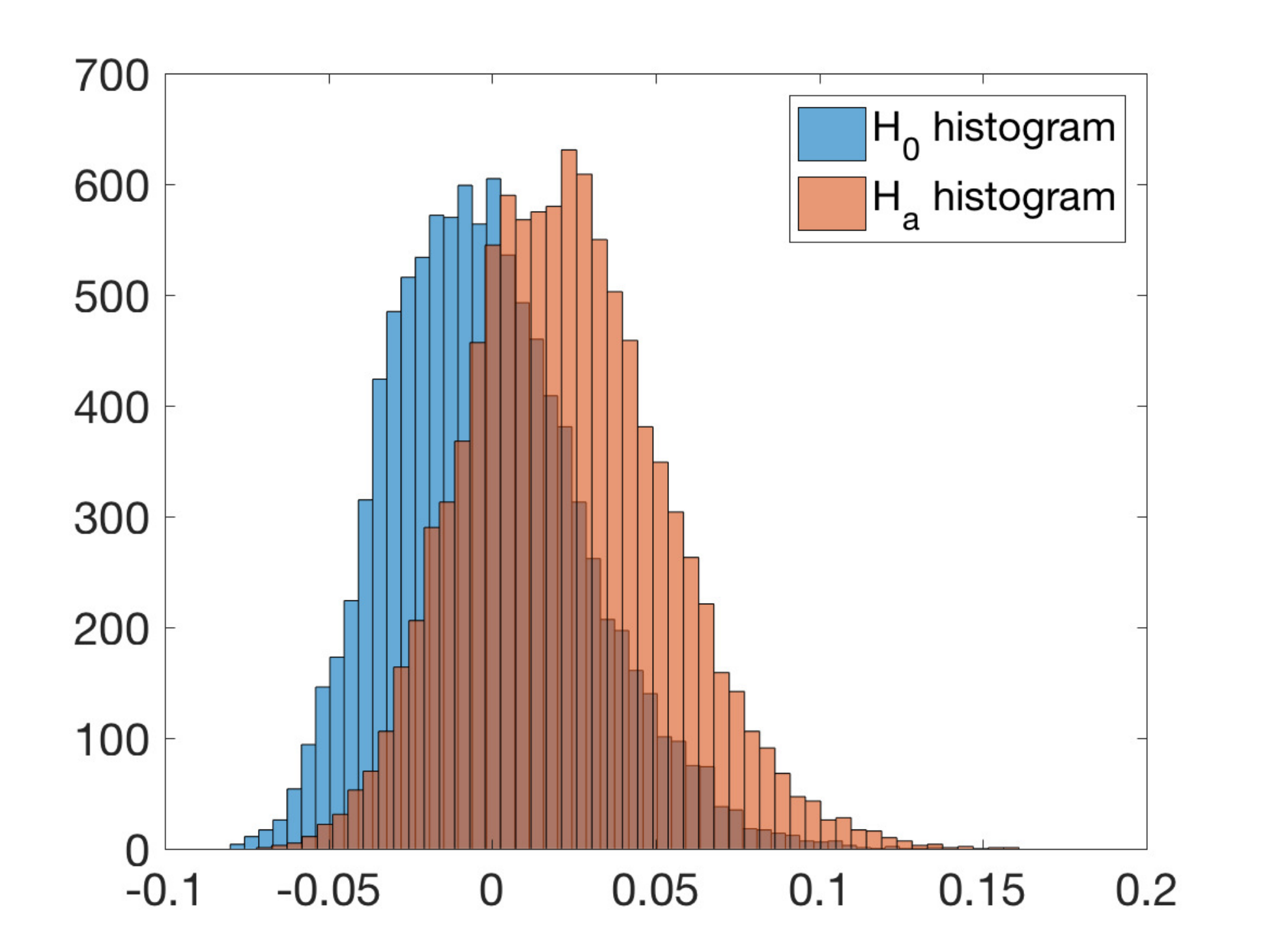}} \\ \vspace{-0.10cm}
(c) MMD$_{b},  B = 20$.
\end{minipage}
\begin{minipage}[t]{0.245\linewidth}
\centering
 {\includegraphics[width=0.99\textwidth]{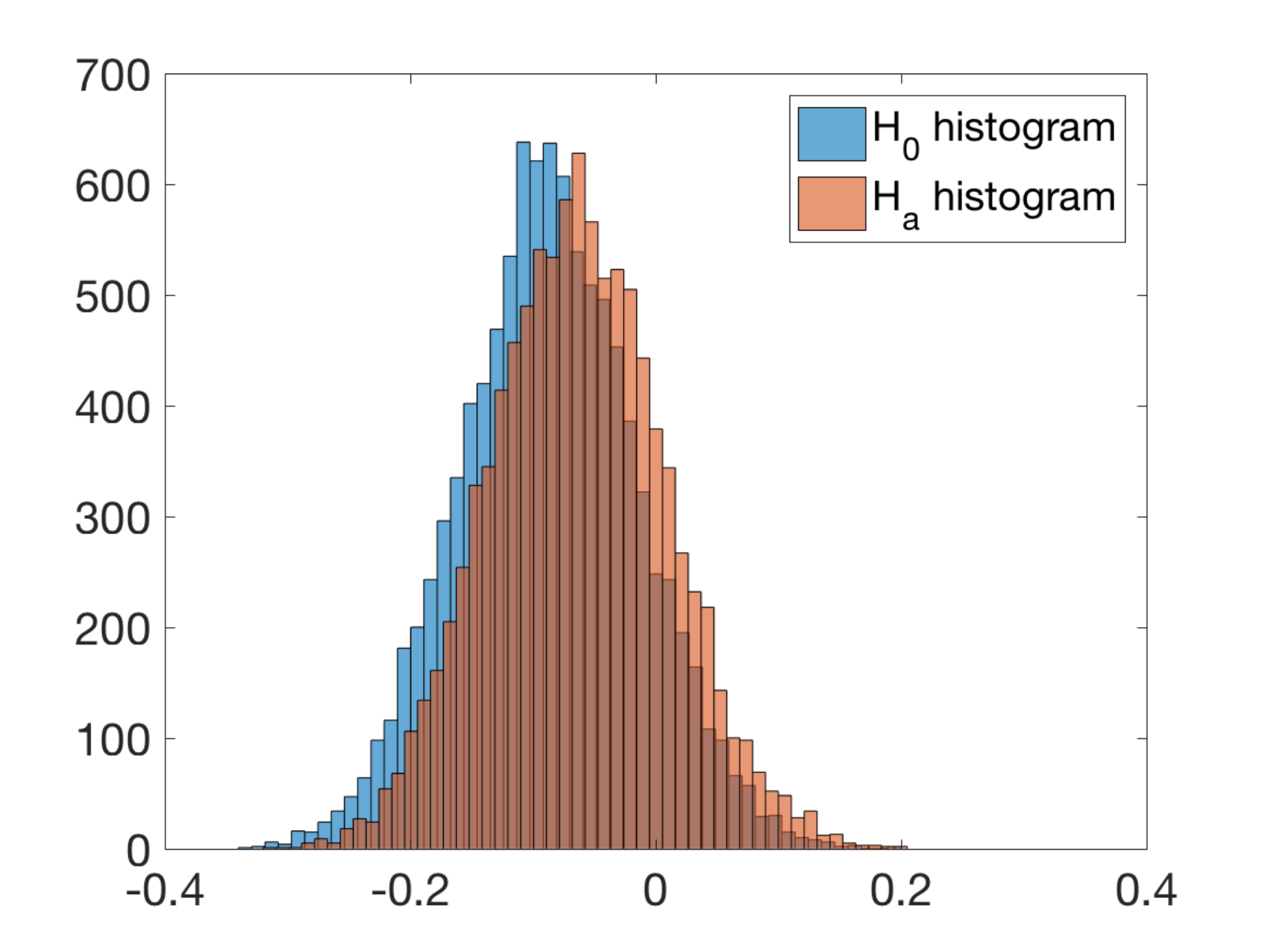}} \\ \vspace{-0.10cm}
(d) MMD$_{b},  B = 5$.
\end{minipage}
\begin{minipage}[t]{0.245\linewidth}
\centering
{\includegraphics[width=0.99\textwidth]{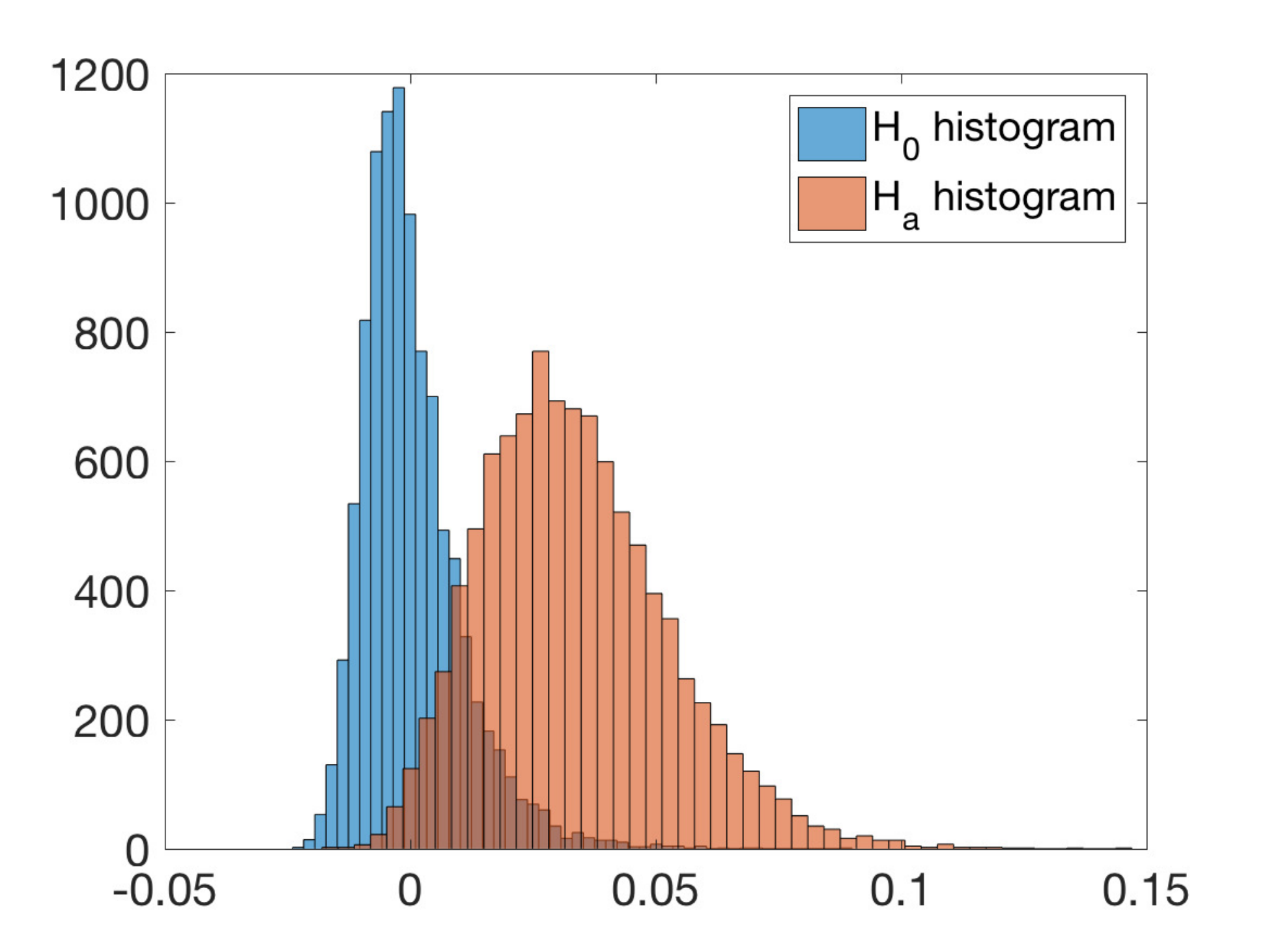}} \\ \vspace{-0.10cm}
(e) MMD$_{inc}, r = 100$.
\end{minipage}
\begin{minipage}[t]{0.245\linewidth}
\centering
 {\includegraphics[width=0.99\textwidth]{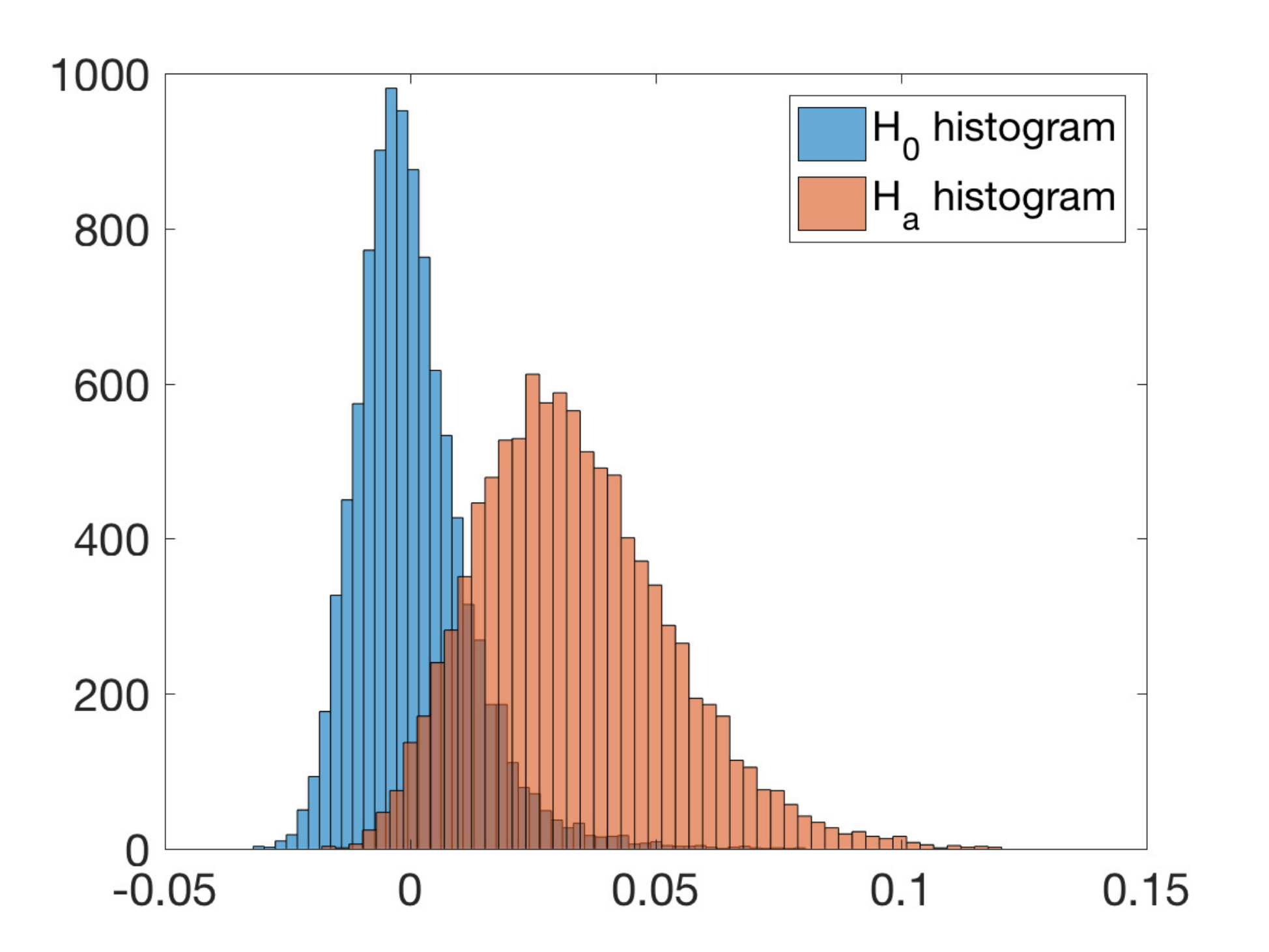}} \\ \vspace{-0.10cm}
(f) MMD$_{inc}, r = 50$.
\end{minipage}
\begin{minipage}[t]{0.245\linewidth}
\centering
{\includegraphics[width=0.99\textwidth]{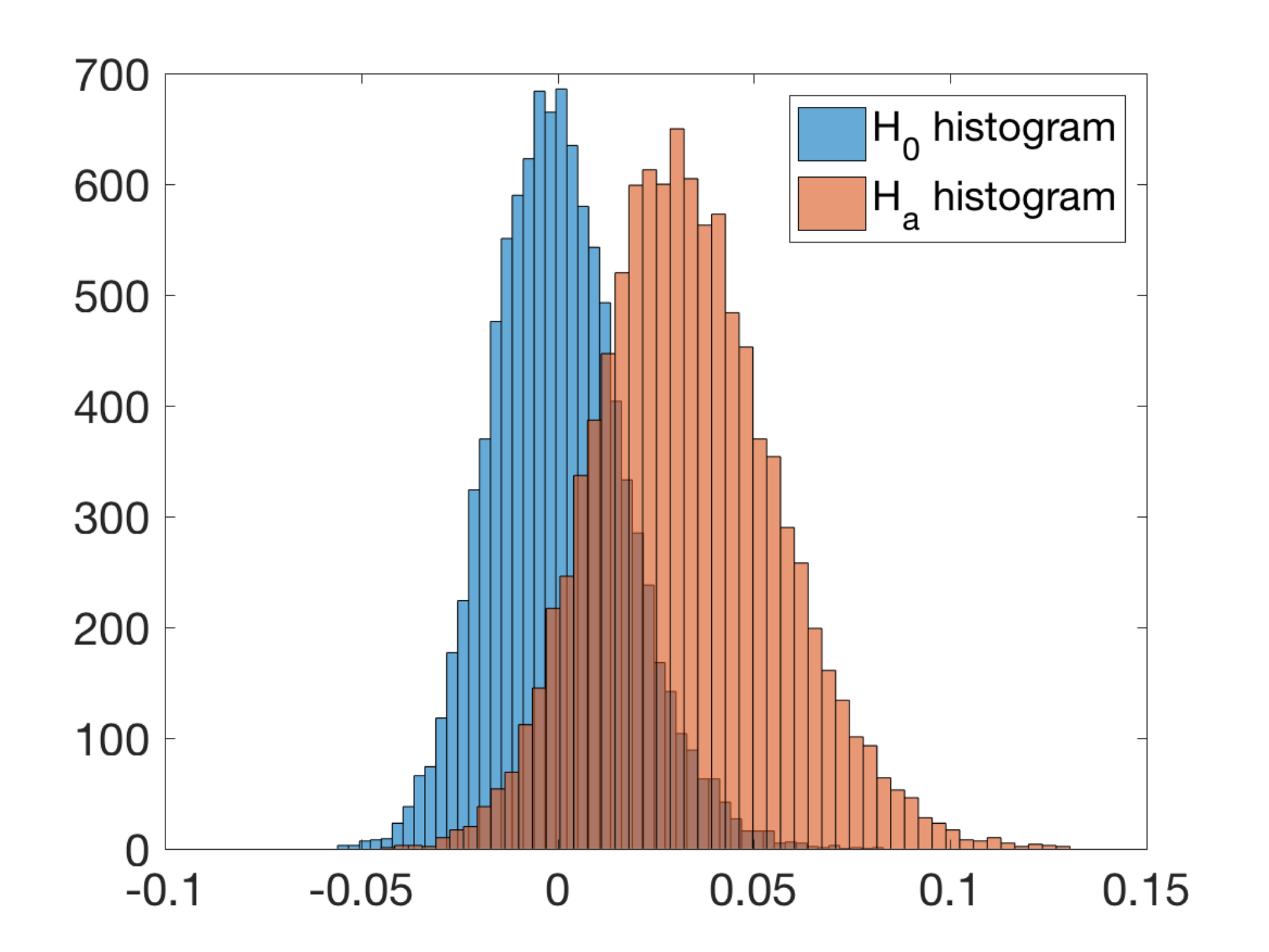}} \\ \vspace{-0.10cm}
(g) MMD$_{inc}, r = 10$.
\end{minipage}
\begin{minipage}[t]{0.245\linewidth}
\centering
 {\includegraphics[width=0.99\textwidth]{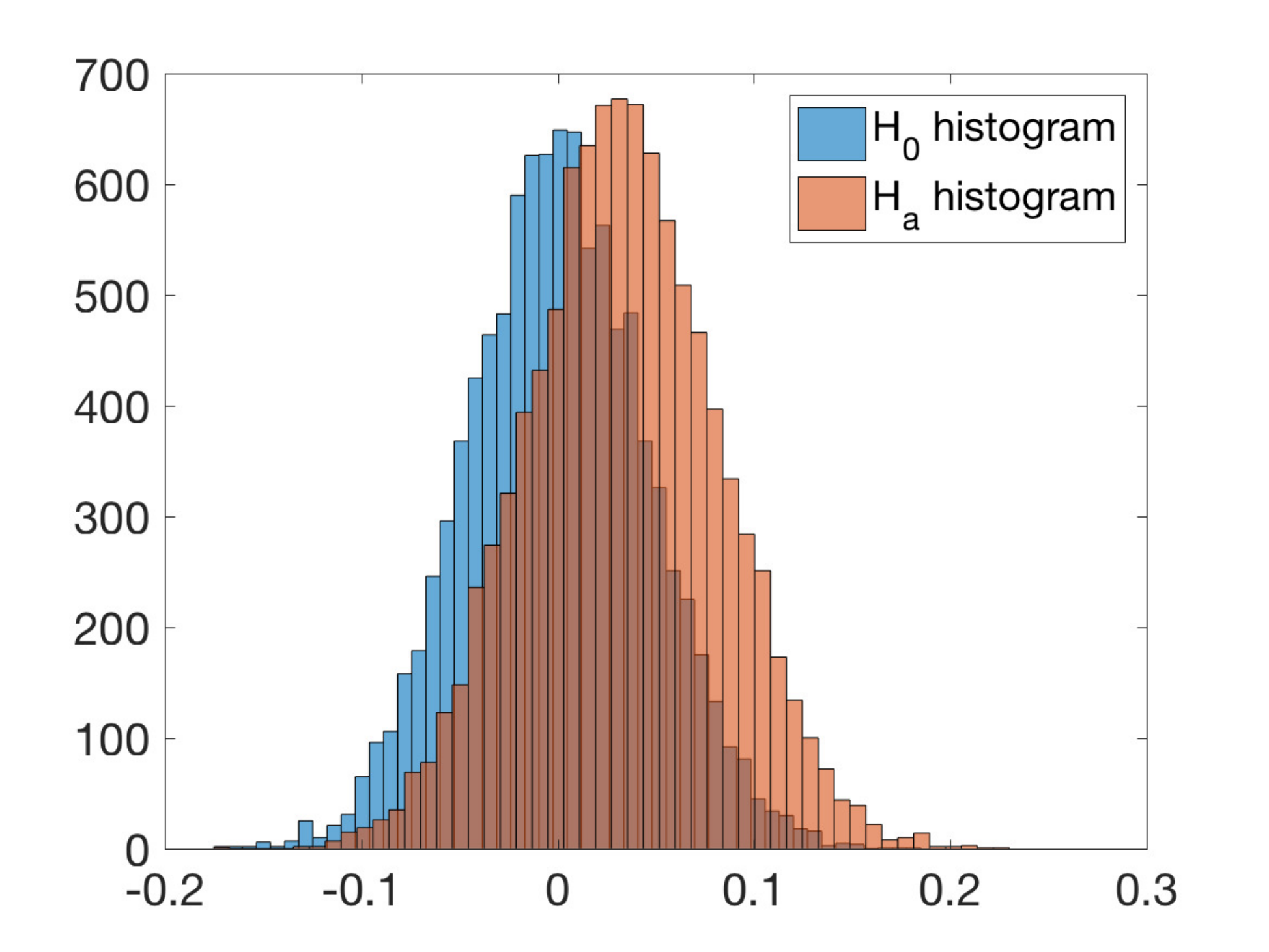}} \\ \vspace{-0.10cm}
(h) MMD$_{inc}, r = 1$.
\end{minipage}
 \caption{Empirical distribution under $p = q$ and $p \neq q$. (a) Complete U-statistics. (b)-(d): The block MMD estimator with different block parameter $B$. (e)-(h): The incomplete MMD estimator with different sampling parameter $r$.  For all plots, we fixed the number of samples as $n = 200$ and the dimensionality $d = 1$.　\label{fig:normality}}
\end{center}
\vspace{-.1in}
\end{figure*}

\begin{figure*}[t!]
\begin{center}
\begin{minipage}[t]{0.245\linewidth}
\centering
{\includegraphics[width=0.99\textwidth]{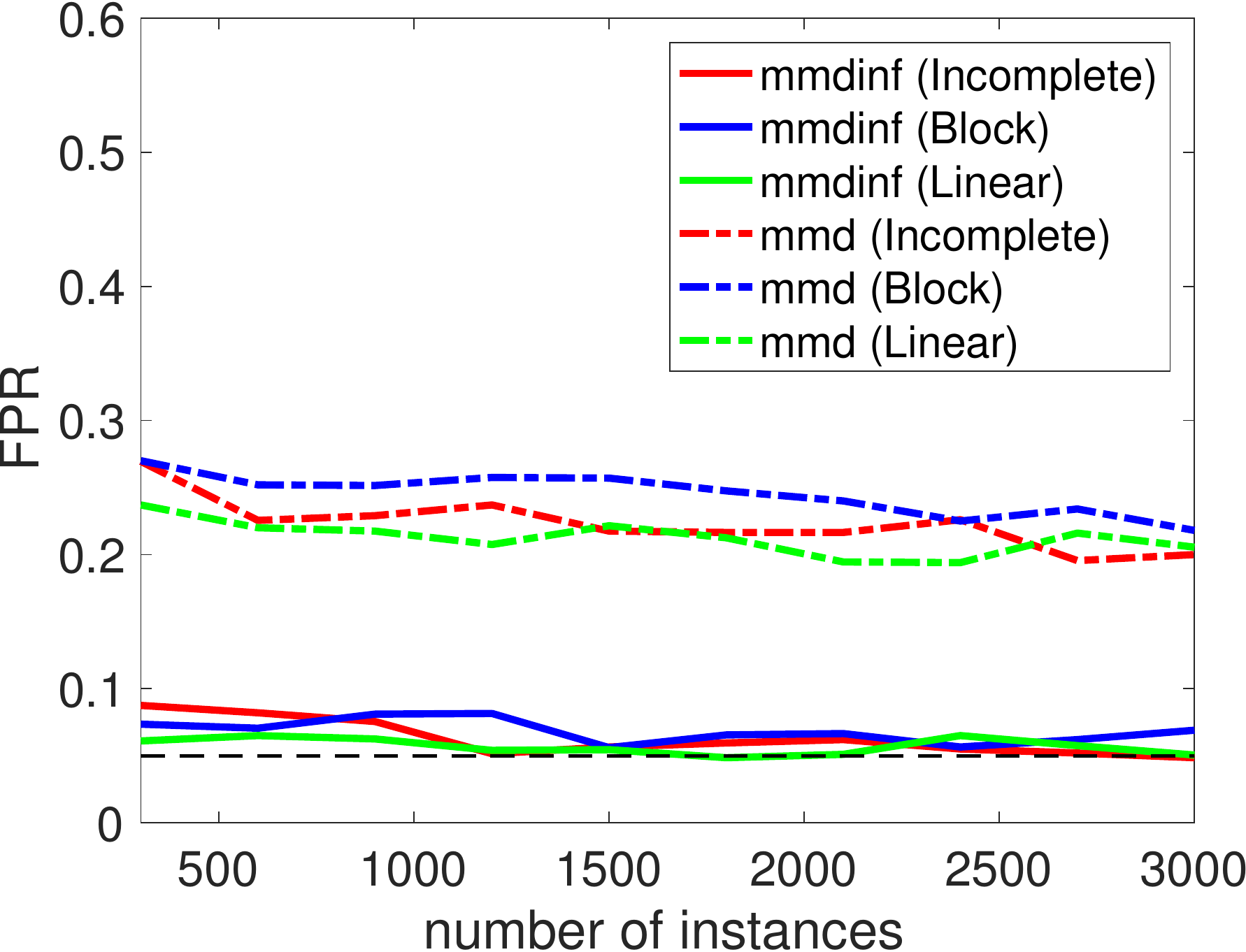}} \\ \vspace{-0.10cm}
(a) Mean shift (FPR).
\end{minipage}
\begin{minipage}[t]{0.245\linewidth}
\centering
 {\includegraphics[width=0.99\textwidth]{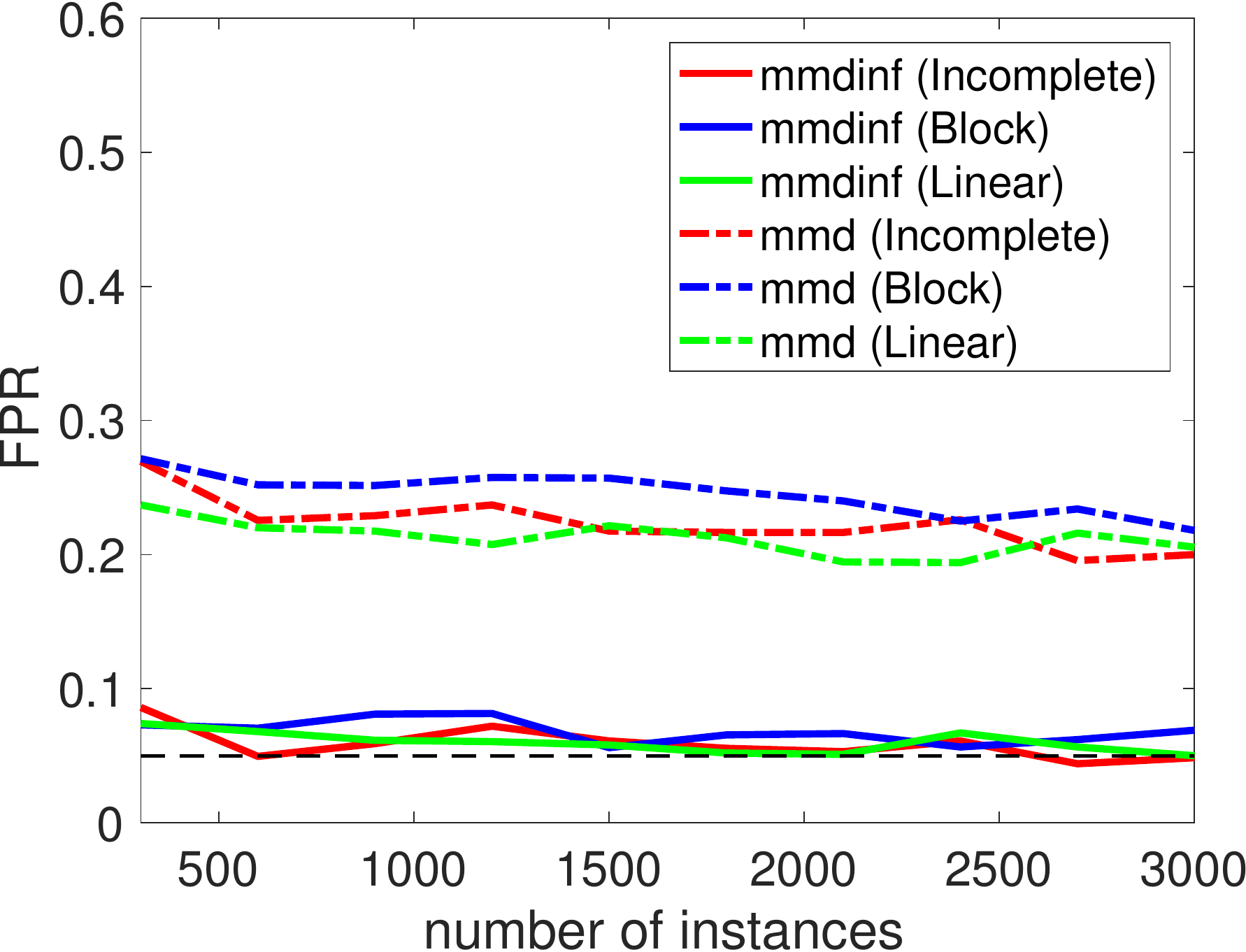}} \\ \vspace{-0.10cm}
(b) Variance shift (FPR).
\end{minipage}
\begin{minipage}[t]{0.245\linewidth}
\centering
{\includegraphics[width=0.99\textwidth]{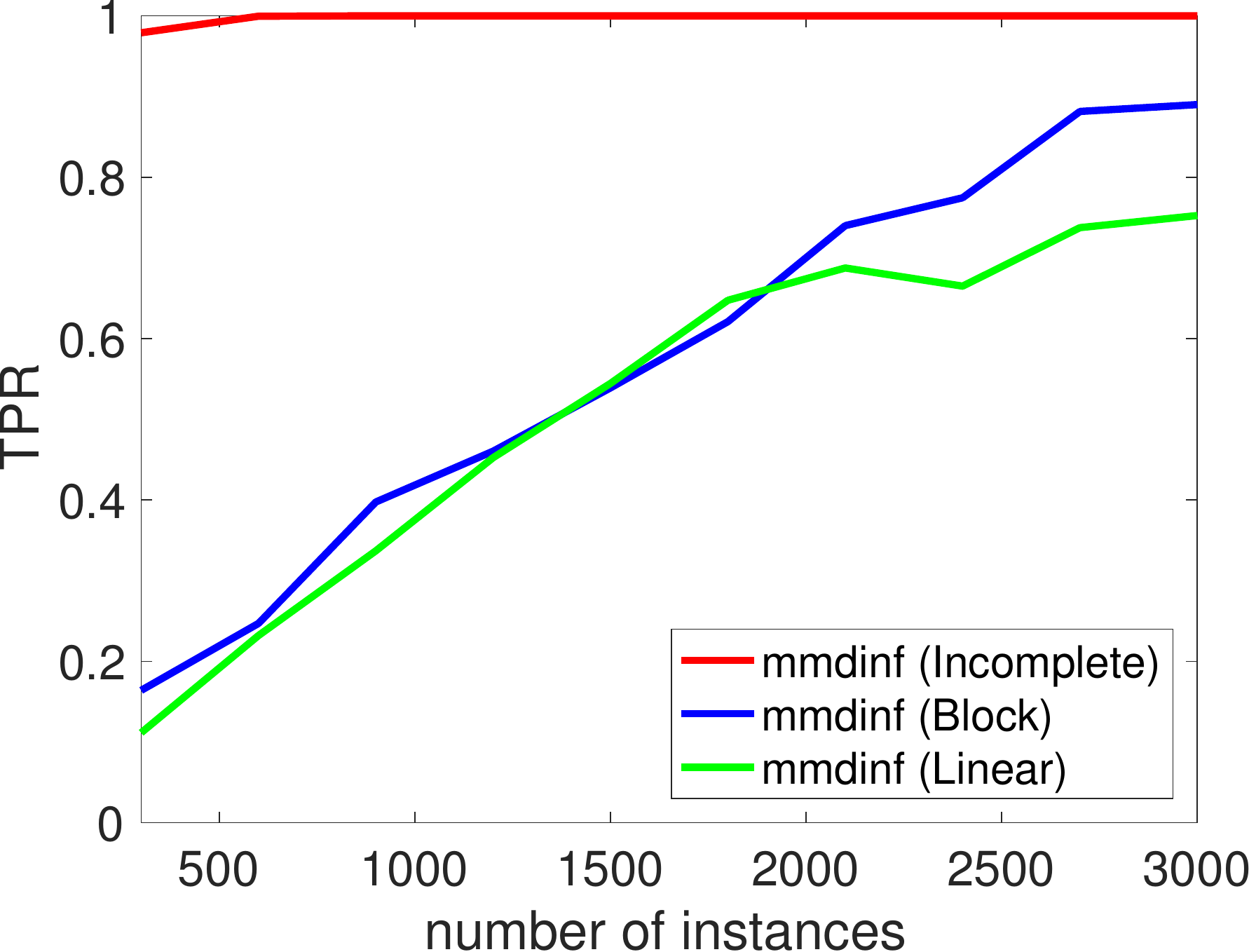}} \\ \vspace{-0.10cm}
(c) Mean shift (TPR).
\end{minipage}
\begin{minipage}[t]{0.245\linewidth}
\centering
 {\includegraphics[width=0.99\textwidth]{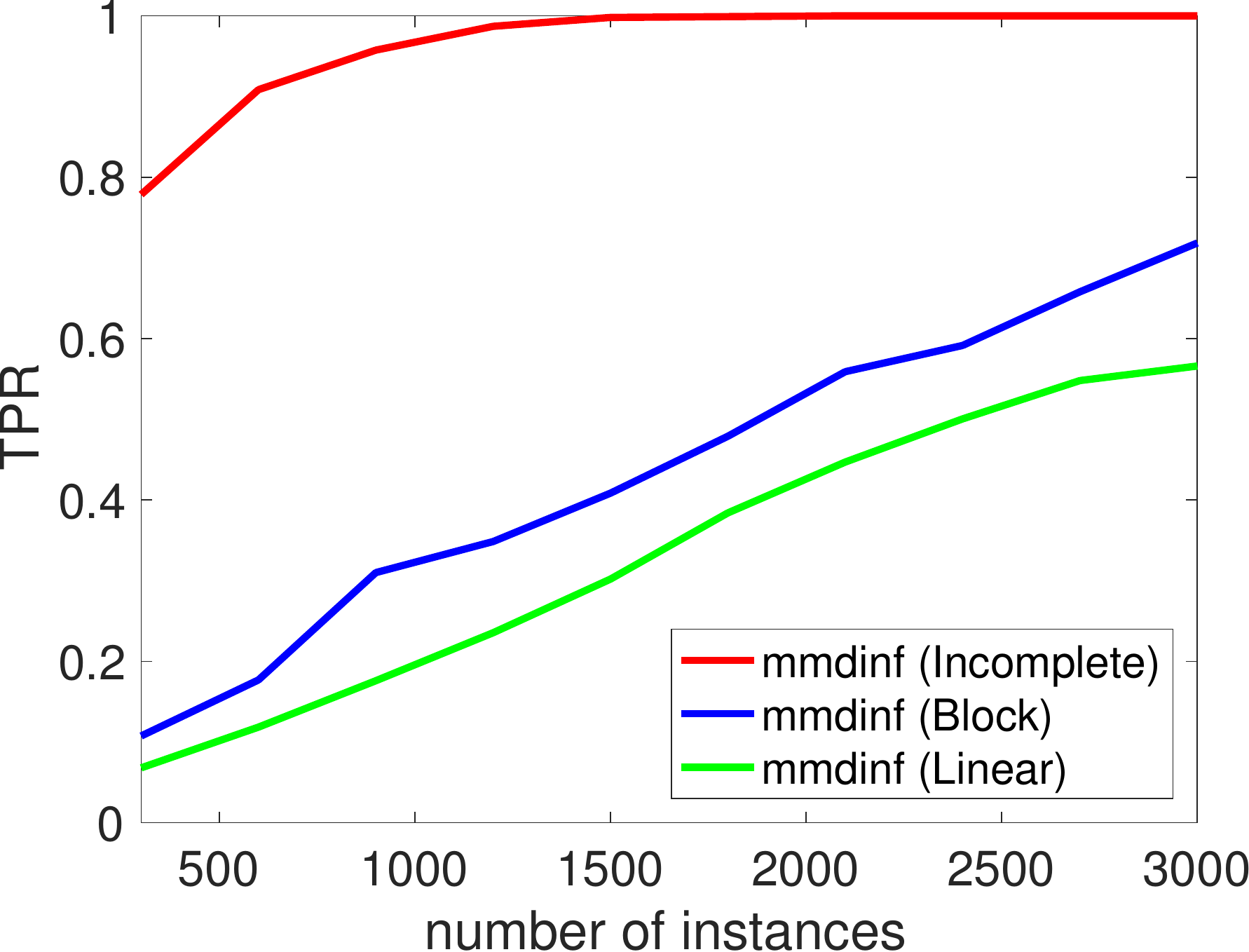}} \\ \vspace{-0.10cm}
(d) Variance shift (TPR).
\end{minipage}
 \caption{(a)(b): False positive rates at significant level $\alpha = 0.05$ of the proposed incomplete estimator, block estimator and linear estimator with/without PSI.  For incomplete MMD, we set $\ell = 10n$. For block MMD, we set the block size $B = \sqrt{n}$. The MMD without PSI computes the $p$-values without adjusting for the selection. (c)(d): True positive rate comparison of the following three empirical estimates for \texttt{mmdInf}. }
    \label{fig:synth1}
\end{center}
\vspace{-.1in}
\end{figure*}

\begin{figure}[ht]
\begin{center}
\begin{minipage}[t]{0.45\linewidth}
\centering
{\includegraphics[width=0.99\textwidth]{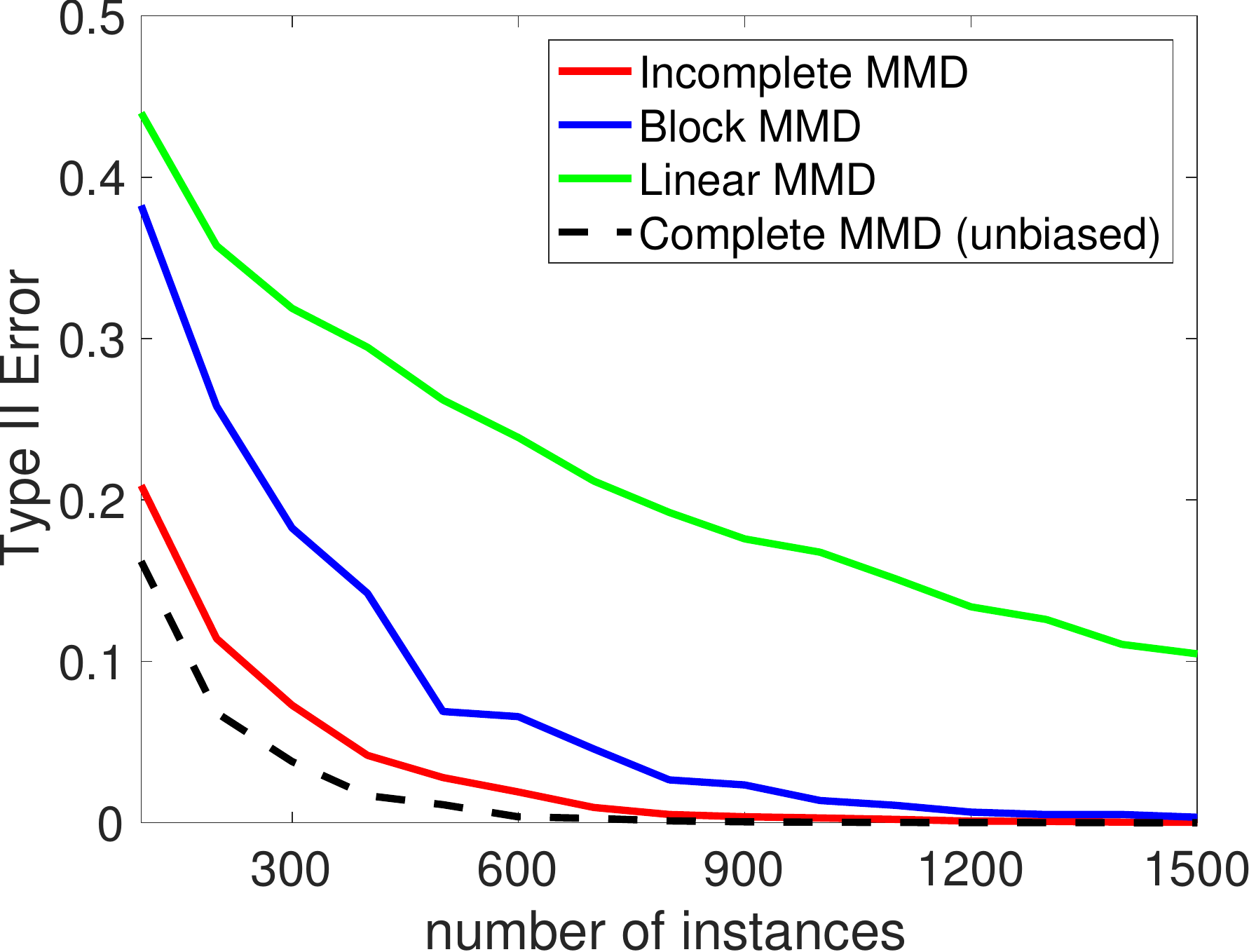}} \\ \vspace{-0.10cm}
(a)
\end{minipage}
\begin{minipage}[t]{0.45\linewidth}
\centering
{\includegraphics[width=0.99\textwidth]{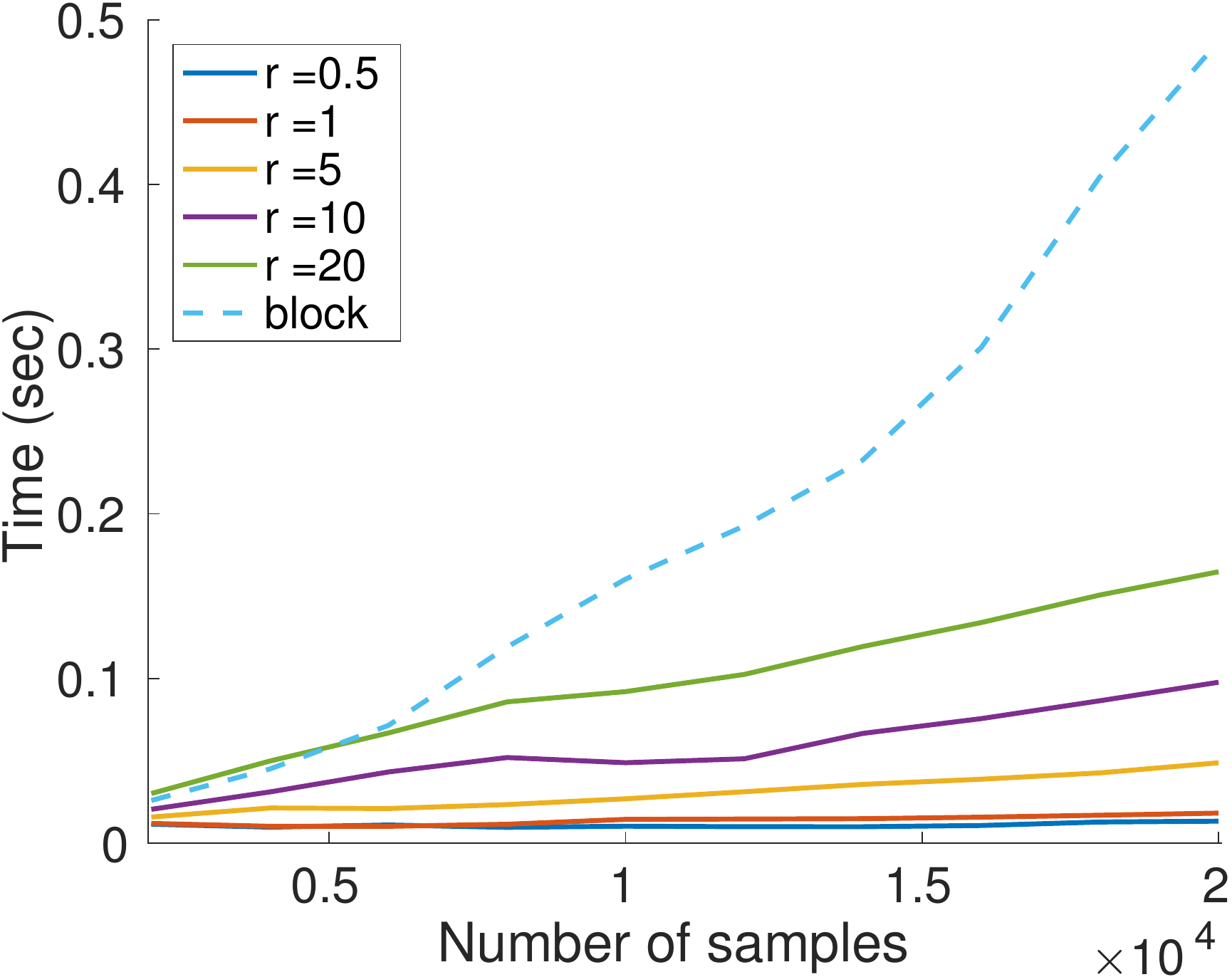}} \\ \vspace{-0.10cm}
(b)
\end{minipage}
 \caption{(a): Type II error comparison. We change the sample size $n = [100, 200 \ldots, 1500]$ and compute the type II error of the four empirical estimates of MMD when the type I error is controlled at 0.05. For incomplete MMD, we use $r = 10$. For the block MMD, we use $B = \sqrt{n}$.  (b): Computational time comparison. We change the sample size $n = [2000, 4000, \ldots, 20000]$ and compute the incomplete MMD and the block MMD, respectively. For incomplete MMD, we use $r = [0.5, 1, 5, 10, 20]$. For the block MMD, we use $B = \sqrt{n}$. Incomplete MMD with $r = 0.5$ (i.e., $\ell = n/2$) can be regarded as the linear-time MMD estimator \citep{gretton2012kernel}. }
    \label{fig:time}
\end{center}
\vspace{-.1in}
\end{figure}

\section{Experiments}
In this section, we evaluate the proposed algorithm on synthetic and real-world datasets. 

\subsection{Setup} We compared \texttt{mmdInf} with a naive testing baseline (\texttt{mmd}), which first selects features using MMD and estimates corresponding $p$-values with the same data of feature selection without adjustment for the selection event. For \texttt{mmdInf}, we used the three MMD estimators: the linear-time MMD \citep{gretton2012kernel}, the block MMD \citep{zaremba2013b}, and the incomplete MMD.  We used $1/3$ of data to calculate the covariance matrix of MMD and the rest to perform feature selection and inference. We fixed the number of selected features (prior to PSI) $k$ to $30$. In PSI the significance of each of the 30 selected features (from ranking MMD) is computed and features with $p$-value lower than the significance level $\alpha = 0.05$ are selected as \emph{statistically significant features}. 
 
For block MMD, in each experiment we set the candidate of block size as $B = \{5, 20, 50\}$ . For incomplete MMD, in each experiment the ratio between number of pairs $(i,j)$ sampled to compute incomplete MMD score and sample size is fixed at $r = \frac{\ell}{n} \in \{0.5,5,10\}$. 

We reported the true positive rate (TPR) $\frac{k'}{k^\ast}$ where $k'$ is the number of true features selected by \texttt{mmdInf} or \texttt{mmd} and $k^\ast$ is the total number of true features in synthetic data. We further computed the false positive rate (FPR) $\frac{k''}{k - k^\ast}$ where $k''$ is the number of \emph{non-true} features reported as positives. We ran all experiments over 200 times with different random seeds, and reported the average TPR and FPR.

\begin{table*}[t!]
\tiny
\centering
\caption{Post selection inference experimental results 
  for real-world datasets. The average TPR and FPR over
  $200$ trials are reported. 
}
\label{tb:benchmark_result}
\begin{tabular}{|@{\ }l@{\ }|c|c||l@{}r|lr|lr|lr|lr|lr|lr|lr|l}
\hline
& 
& 
&\multicolumn{2}{c|}{
Linear-Time
} 
& \multicolumn{6}{c|}{
Block 
}
& \multicolumn{6}{c|}{
Incomplete 
} \\
Datasets  
& $d$
& $n$
&\multicolumn{2}{c|}{
} 
& \multicolumn{2}{c|}{
$B = 5$
}
& \multicolumn{2}{c|}{
$B = 20$
} 
& \multicolumn{2}{c|}{
$B = 50$
} 
& \multicolumn{2}{c|}{
$r = 0.5$
} 
& \multicolumn{2}{c|}{
$r = 5$
} 
& \multicolumn{2}{c|}{
$r = 10$
} \\ 
&  & & TPR & FPR & TPR  & FPR  & TPR  & FPR & TPR & FPR& TPR & FPR & TPR & FPR  & TPR & FPR \\ \hline
Diabetis       & 8   & 768  & 0.05 & 0.05  & 0.06 & 0.05 & 0.12 & 0.06 & 0.21 & 0.11  & 0.07 & 0.07  & 0.41 & 0.07 & 0.55 & 0.07 \\ \hline
Wine (Red)   & 11  & 4898 & 0.15 & 0.06 & 0.14 & 0.02  & 0.29 & 0.05 & 0.33 & 0.07 &  0.21 &  0.06& 0.64 & 0.06 & 0.74 & 0.06\\ \hline
Wine (White) & 11 & 1599 & 0.09 & 0.06  & 0.09 & 0.05 & 0.13 & 0.06 & 0.18 & 0.06 & 0.09 & 0.06 & 0.43  & 0.06 & 0.53 & 0.06\\ \hline
Australia      & 7 & 690     & 0.08 & 0.06  & 0.08 & 0.05 & 0.15 & 0.07 & 0.39 & 0.23 & 0.08 & 0.05 &  0.52 & 0.06 & 0.69 & 0.07 \\ \hline
\end{tabular}
\vspace{-.1in}
\end{table*}

\subsection{Illustrative experiments}
Figure~\ref{fig:normality} shows the empirical distribution under $p = q$ and $p \neq q$ for the complete estimator, the block estimator and the incomplete estimator. As can be observed, the empirical distribution of the incomplete estimator is normal for small sampling parameter $r$, and becomes similar to its complete counterpart if $r$ is large; this is supported by Theorem~\ref{theo_mmd_normal1} ($\gamma = \infty$). Moreover, compared to the block estimator, the incomplete estimator tends to have a better trade-off between variance and normality. 

Figure~\ref{fig:time}(a) shows the Type II error comparison for two-sample test with one dimensional Gaussian mean shift data. The Type II error is computed when the Type I error is fixed at 0.05, and the incomplete MMD outperforms other estimators. Figure~\ref{fig:time} (b) compares the computational time of the empirical estimates, and for small $r$ the computational time of incomplete MMD is much less than that of the block MMD. Overall, the incomplete MMD has favorable properties in practice.

\subsection{Synthetic experiments (PSI)}
The number  of features $d$ is fixed to $50$, and for each feature, data is randomly generated following a Gaussian distribution with set mean and variance. $10$ out of the 50 features are set to be significantly different by shifting the distribution of one class away from the other (mean or variance).   More specifically, we generate the synthetic data as
\begin{description}
\item[(a) Mean shift] $\boldx \sim \calN(\boldzero_{50}, \boldI_{50})$, $\boldy \sim \calN(\boldmu, \boldI)$, $\boldmu = [\boldone_{10}^\top ~ \boldzero_{40}^\top]^\top \in \mathbbR^{50}$,
\item[(b) Variance shift] $\boldx \sim \calN(\boldzero_{50}, \boldI_{50})$, $\boldy \sim \calN(\boldzero, \boldSigma)$, $\boldSigma = \text{diag}([1.5 \boldone_{10}^\top~ \boldone_{40}^\top]^\top)$
\end{description}
where $\calN(\boldmu, \boldSigma)$ is a multivariate normal distribution with mean $\boldmu \in \mathbbR^{d}$ and covariance $\boldSigma \in \mathbbR^{d\times d}$, $\boldone_p \in \mathbbR^d$ is a $d$-dimensional vector whose elements are all one,  $\boldzero_d \in \mathbbR^d$ is a $d$-dimensional vector whose elements are all zero, and $\text{diag}(\bolda) \in \mathbbR^{d\times d}$ is a diagonal matrix whose diagonal elements are $\bolda \in \mathbbR^d$.

\vspace{.05in}
\noindent {\bf False positive rate control:} Figure~\ref{fig:synth1}(a) and (b) show the FPRs of linear MMD, block MMD and incomplete MMD with or without PSI. As can be clearly seen, PSI successfully controls FPR with significance level $\alpha = 0.05$ for all the three estimators, whereas the naive approach tends to have higher FPRs.

\vspace{.05in}
\noindent {\bf True positive rate comparison:} Figures~\ref{fig:synth1} shows the TPRs of the synthetic data. In both cases, the TPR of incomplete MMD converges to 1 significantly faster than the the other two empirical estimates.

\subsection{Real-world data (Benchmark)}
We compared the proposed algorithm by using real-world datasets. 
Since it is difficult to decide what is a "true feature" in real-world data, we choose a few datasets for binary classification with small amount of features, and regard all the original features as \emph{true}. We then concatenated random features to the \emph{true} features  (the total number of features $d = 100$). Table~\ref{tb:benchmark_result} shows TPRs and FPRs of \texttt{mmdInf} with different MMD estimators. It can be observed that the incomplete estimator significantly outperforms the other empirical estimates. Note that a higher TPR can be achieved with higher $r$, while the FPR is still controlled at 0.05 with the highest $r = 10$ that we chose. 

\subsection{GANs analysis (Sample selection)}
We also applied \texttt{mmdInf} for evaluating the generation quality of GANs. We trained BEGAN \citep{berthelot2017began}, DCGAN \citep{radford2015unsupervised}, STDGAN \citep{miyato2017spectral}, Cramer GAN \citep{bellemare2017cramer}, DFM \citep{warde2016improving}, DRAGAN \citep{kodali2017convergence}, and Minibatch Discrimination GAN \citep{salimans2016improved}, generated 5000 images (using Chainer GAN package \footnote{\url{https://github.com/pfnet-research/chainer-gan-lib}} with CIFAR10 datasets), and extracted 512 dimensional features by pre-trained Resnet18 \citep{he2016deep}. For the true image sets, we subsampled $5000$ images from CIFAR10 datasets and computed the 512 dimensional features using the same Resnet18. We then tested the difference between the generated images and the real images using \texttt{mmdInf} on the extracted features (see Sec. 3.4).

\begin{figure}[t!]
\begin{center}
\begin{minipage}[t]{0.45\linewidth}
\centering
{\includegraphics[width=0.99\textwidth]{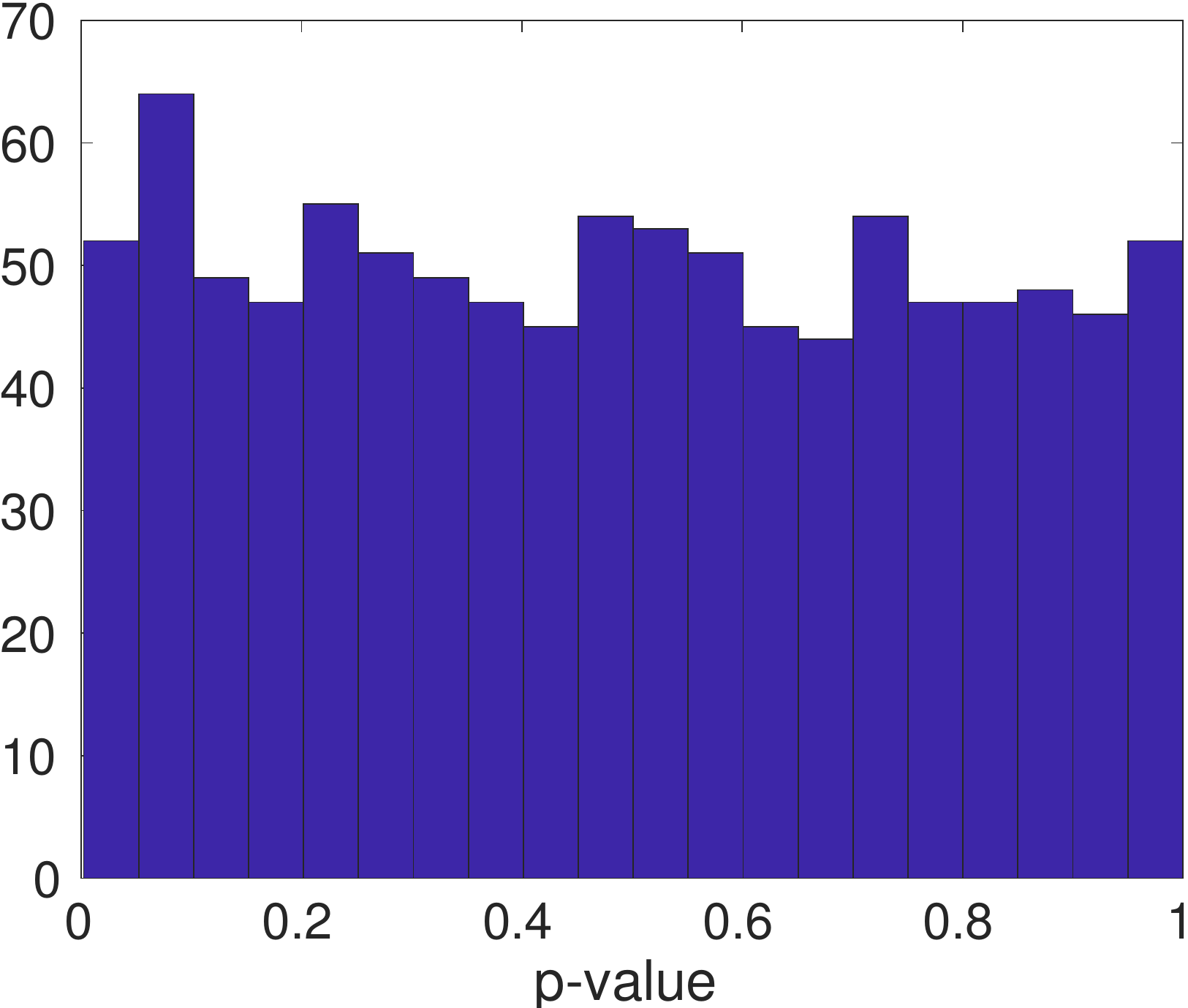}} \\ \vspace{-0.10cm}
(a) 
\end{minipage}
\begin{minipage}[t]{0.45\linewidth}
\centering
{\includegraphics[width=0.99\textwidth]{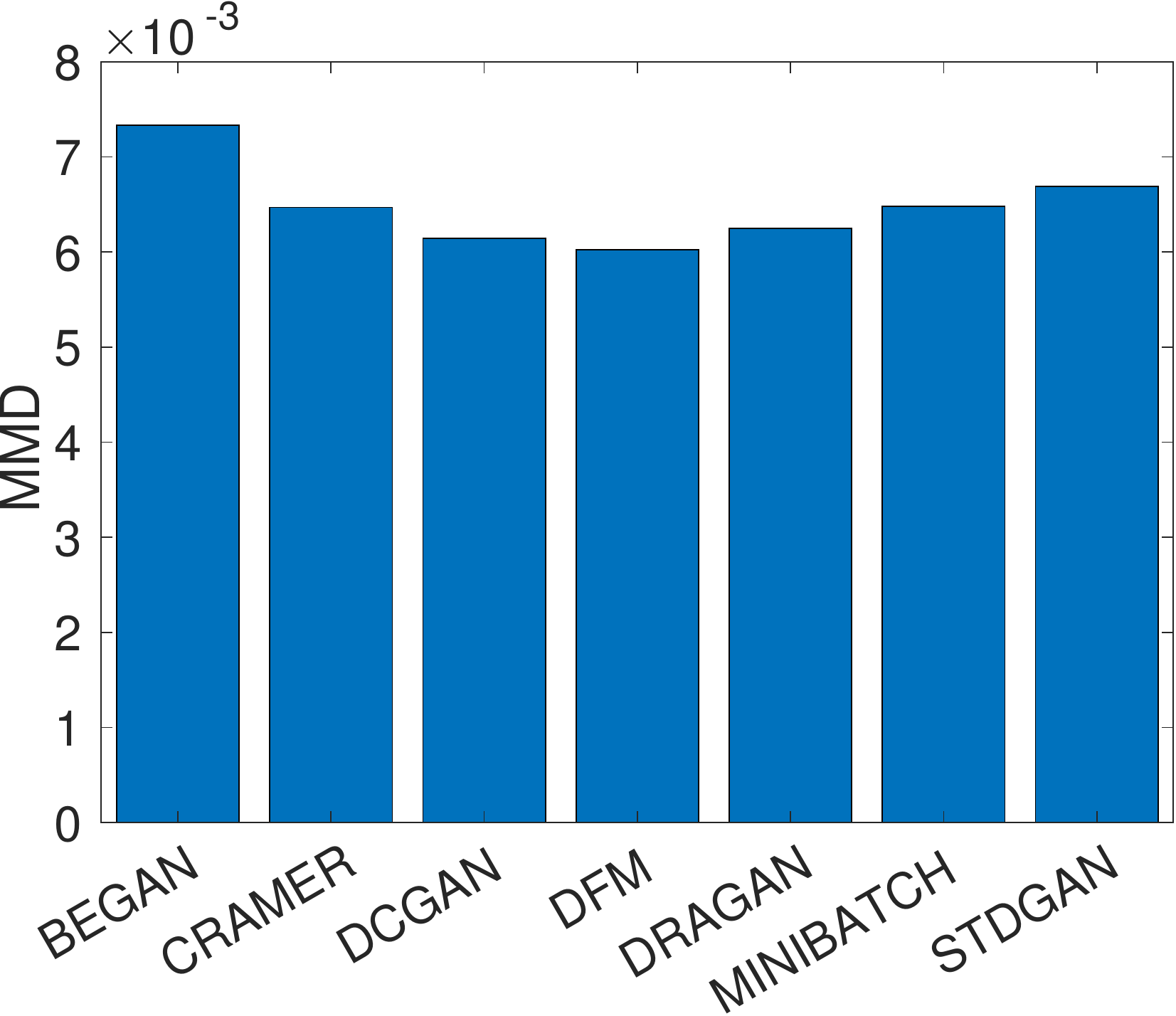}} \\ \vspace{-0.10cm}
(b) 
\end{minipage}
 \caption{(a) Histogram of $p$-values over 1000 runs. (b) Averaged incomplete MMD scores. }
    \label{fig:gan-pval}
\end{center}
\vspace{-.1in}
\end{figure}

However, we found that for all the members in the GAN family, the null hypothesis was rejected, i.e., the generated distribution and the real distribution are different. As sanity check, we evaluated \texttt{mmdInf} by constructing an "oracle" generative model that generates real images from CIFAR10. Next, we randomly selected 5000 images (a disjoint set from the oracle generative images) from CIFAR10 in each trial, and set the sampling ratio to $r = 5$. Figure~\ref{fig:gan-pval}(a) shows the distribution of $p$-values computed by our algorithm. We can see that the $p$-values are distributed uniformly in the tests for the "oracle" generative model, which matches the theoretical result in Theorem~\ref{theo-psi}. Thus the algorithm is able to detect the distribution difference and control the false positive rate. In other words, if the generated GANs samples do not follow the original distribution, we can safely reject the null hypothesis with a given significance level $\alpha$. 

Figure~\ref{fig:gan-pval}(b) shows the estimated MMD scores of each member in GANs family. Based on the results, we could tell that DFM was the best model and DCGAN was the second best model to generate images following the true distribution. However, the difference between various members is not obvious. Developing a validation pipeline based on \texttt{mmdInf} for GANs analysis would be one interesting line of future work.

\section{Conclusion}
In this paper, we proposed a novel statistical testing framework \texttt{mmdInf}, which can find a set of \emph{statistically significant} features that can discriminate two distributions. Through synthetic and real-world experiments, we demonstrated that \texttt{mmdInf} can successfully find important features and/or datasets. We also proposed a method for sample selection based on \texttt{mmdInf} and applied it in the evaluation of generative models.

\section{Acknowledgement}
We would like to thank Dr. Kazuki Yoshizoe, Prof. Yuta Umezu , and Prof. Suriya Gunasekar for fruitful discussions and valuable suggestions. MY was supported by the JST PRESTO program JPMJPR165A and partly supported by MEXT KAKENHI 16K16114.

\bibliography{main}

\begin{thebibliography}{42}
\providecommand{\natexlab}[1]{#1}
\providecommand{\url}[1]{\texttt{#1}}
\expandafter\ifx\csname urlstyle\endcsname\relax
  \providecommand{\doi}[1]{doi: #1}\else
  \providecommand{\doi}{doi: \begingroup \urlstyle{rm}\Url}\fi

\bibitem[Ali and Silvey(1966)]{ali1966general}
Syed~Mumtaz Ali and Samuel~D Silvey.
\newblock A general class of coefficients of divergence of one distribution
  from another.
\newblock \emph{Journal of the Royal Statistical Society. Series B
  (Methodological)}, pages 131--142, 1966.

\bibitem[Anderson(2001)]{anderson2001new}
Marti~J Anderson.
\newblock A new method for non-parametric multivariate analysis of variance.
\newblock \emph{Austral ecology}, 26\penalty0 (1):\penalty0 32--46, 2001.

\bibitem[Bellemare et~al.(2017)Bellemare, Danihelka, Dabney, Mohamed,
  Lakshminarayanan, Hoyer, and Munos]{bellemare2017cramer}
Marc~G Bellemare, Ivo Danihelka, Will Dabney, Shakir Mohamed, Balaji
  Lakshminarayanan, Stephan Hoyer, and R{\'e}mi Munos.
\newblock The cramer distance as a solution to biased wasserstein gradients.
\newblock \emph{arXiv preprint arXiv:1705.10743}, 2017.

\bibitem[Berthelot et~al.(2017)Berthelot, Schumm, and Metz]{berthelot2017began}
David Berthelot, Tom Schumm, and Luke Metz.
\newblock Began: Boundary equilibrium generative adversarial networks.
\newblock \emph{arXiv preprint arXiv:1703.10717}, 2017.

\bibitem[Blom(1976)]{blom1976some}
Gunnar Blom.
\newblock Some properties of incomplete u-statistics.
\newblock \emph{Biometrika}, 63\penalty0 (3):\penalty0 573--580, 1976.

\bibitem[Che et~al.(2016)Che, Li, Jacob, Bengio, and Li]{che2016mode}
Tong Che, Yanran Li, Athul~Paul Jacob, Yoshua Bengio, and Wenjie Li.
\newblock Mode regularized generative adversarial networks.
\newblock \emph{arXiv preprint arXiv:1612.02136}, 2016.

\bibitem[Cover and Thomas(2012)]{cover2012elements}
Thomas~M Cover and Joy~A Thomas.
\newblock \emph{Elements of information theory}.
\newblock John Wiley \& Sons, 2012.

\bibitem[Friedman et~al.(2001)Friedman, Hastie, and
  Tibshirani]{friedman2001elements}
Jerome Friedman, Trevor Hastie, and Robert Tibshirani.
\newblock \emph{The elements of statistical learning}, volume~1.
\newblock Springer series in statistics New York, 2001.

\bibitem[Goodfellow et~al.(2014)Goodfellow, Pouget-Abadie, Mirza, Xu,
  Warde-Farley, Ozair, Courville, and Bengio]{goodfellow2014generative}
Ian Goodfellow, Jean Pouget-Abadie, Mehdi Mirza, Bing Xu, David Warde-Farley,
  Sherjil Ozair, Aaron Courville, and Yoshua Bengio.
\newblock Generative adversarial nets.
\newblock In \emph{NIPS}, 2014.

\bibitem[Gretton et~al.(2005)Gretton, Bousquet, Smola, and
  Scholkopf]{gretton2005measuring}
Arthur Gretton, Olivier Bousquet, Alex Smola, and Bernhard Scholkopf.
\newblock Measuring statistical dependence with hilbert-schmidt norms.
\newblock In \emph{ALT}, 2005.

\bibitem[Gretton et~al.(2007)Gretton, Borgwardt, Rasch, Sch{\"o}lkopf, and
  Smola]{gretton2007kernel}
Arthur Gretton, Karsten~M Borgwardt, Malte Rasch, Bernhard Sch{\"o}lkopf, and
  Alex~J Smola.
\newblock A kernel method for the two-sample-problem.
\newblock In \emph{NIPS}, 2007.

\bibitem[Gretton et~al.(2012)Gretton, Borgwardt, Rasch, Sch{\"o}lkopf, and
  Smola]{gretton2012kernel}
Arthur Gretton, Karsten~M Borgwardt, Malte~J Rasch, Bernhard Sch{\"o}lkopf, and
  Alexander Smola.
\newblock A kernel two-sample test.
\newblock \emph{JMLR}, 13\penalty0 (Mar):\penalty0 723--773, 2012.

\bibitem[He et~al.(2016)He, Zhang, Ren, and Sun]{he2016deep}
Kaiming He, Xiangyu Zhang, Shaoqing Ren, and Jian Sun.
\newblock Deep residual learning for image recognition.
\newblock In \emph{CVPR}, 2016.

\bibitem[Heusel et~al.(2017)Heusel, Ramsauer, Unterthiner, Nessler, Klambauer,
  and Hochreiter]{heusel2017gans}
Martin Heusel, Hubert Ramsauer, Thomas Unterthiner, Bernhard Nessler,
  G{\"u}nter Klambauer, and Sepp Hochreiter.
\newblock Gans trained by a two time-scale update rule converge to a nash
  equilibrium.
\newblock \emph{arXiv preprint arXiv:1706.08500}, 2017.

\bibitem[Huang et~al.(2018)Huang, Yuan, Xu, Guo, Sun, Wu, and
  Weinberge]{huang2018an}
Gao Huang, Yang Yuan, Qiantong Xu, Chuan Guo, Yu~Sun, Felix Wu, and Kilian
  Weinberge.
\newblock An empirical study on evaluation metrics of generative adversarial
  networks.
\newblock \emph{arXiv preprint arXiv:1610.06545}, 2018.

\bibitem[Janson(1984)]{janson1984asymptotic}
Svante Janson.
\newblock The asymptotic distributions of incomplete u-statistics.
\newblock \emph{Probability Theory and Related Fields}, 66\penalty0
  (4):\penalty0 495--505, 1984.

\bibitem[Jitkrittum et~al.(2016)Jitkrittum, Szab{\'o}, Chwialkowski, and
  Gretton]{jitkrittum2016interpretable}
Wittawat Jitkrittum, Zolt{\'a}n Szab{\'o}, Kacper~P Chwialkowski, and Arthur
  Gretton.
\newblock Interpretable distribution features with maximum testing power.
\newblock In \emph{NIPS}, 2016.

\bibitem[Kanamori et~al.(2009)Kanamori, Hido, and Sugiyama]{kanamori2009least}
Takafumi Kanamori, Shohei Hido, and Masashi Sugiyama.
\newblock A least-squares approach to direct importance estimation.
\newblock \emph{JMLR}, 10\penalty0 (Jul):\penalty0 1391--1445, 2009.

\bibitem[Kodali et~al.(2017)Kodali, Abernethy, Hays, and
  Kira]{kodali2017convergence}
Naveen Kodali, Jacob Abernethy, James Hays, and Zsolt Kira.
\newblock On convergence and stability of gans.
\newblock \emph{arXiv preprint arXiv:1705.07215}, 2017.

\bibitem[Lander and Kruglyak(1995)]{lander1995genetic}
Eric Lander and Leonid Kruglyak.
\newblock Genetic dissection of complex traits: guidelines for interpreting and
  reporting linkage results.
\newblock \emph{Nature genetics}, 11\penalty0 (3):\penalty0 241, 1995.

\bibitem[Lee et~al.(2016)Lee, Sun, Sun, Taylor, et~al.]{lee2016exact}
Jason~D Lee, Dennis~L Sun, Yuekai Sun, Jonathan~E Taylor, et~al.
\newblock Exact post-selection inference, with application to the lasso.
\newblock \emph{The Annals of Statistics}, 44\penalty0 (3):\penalty0 907--927,
  2016.

\bibitem[Lee(1990)]{lee1990u}
Justin Lee.
\newblock U-statistics: Theory and practice.
\newblock 1990.

\bibitem[Li et~al.(2015{\natexlab{a}})Li, Xie, Dai, and Song]{li2015m}
Shuang Li, Yao Xie, Hanjun Dai, and Le~Song.
\newblock M-statistic for kernel change-point detection.
\newblock In \emph{NIPS}, 2015{\natexlab{a}}.

\bibitem[Li et~al.(2015{\natexlab{b}})Li, Swersky, and Zemel]{Li2015mmd}
Yujia Li, Kevin Swersky, and Richard Zemel.
\newblock Generative moment matching networks.
\newblock In \emph{NIPS}, 2015{\natexlab{b}}.

\bibitem[Liu et~al.(2013)Liu, Yamada, Collier, and Sugiyama]{liu2013change}
Song Liu, Makoto Yamada, Nigel Collier, and Masashi Sugiyama.
\newblock Change-point detection in time-series data by relative density-ratio
  estimation.
\newblock \emph{Neural Networks}, 43:\penalty0 72--83, 2013.

\bibitem[Lopez-Paz and Oquab(2016)]{lopez2016revisiting}
David Lopez-Paz and Maxime Oquab.
\newblock Revisiting classifier two-sample tests.
\newblock \emph{arXiv preprint arXiv:1610.06545}, 2016.

\bibitem[Miyato et~al.(2017)Miyato, Kataoka, Koyama, and
  Yoshida]{miyato2017spectral}
Takeru Miyato, Toshiki Kataoka, Masanori Koyama, and Yuichi Yoshida.
\newblock Spectral normalization for generative adversarial networks.
\newblock In \emph{ICML Implicit Models Workshop}, 2017.

\bibitem[Mueller and Jaakkola(2015)]{mueller2015principal}
Jonas~W Mueller and Tommi Jaakkola.
\newblock Principal differences analysis: Interpretable characterization of
  differences between distributions.
\newblock In \emph{NIPS}, 2015.

\bibitem[Nowozin et~al.(2016)Nowozin, Cseke, and Tomioka]{nowozin2016f}
Sebastian Nowozin, Botond Cseke, and Ryota Tomioka.
\newblock f-gan: Training generative neural samplers using variational
  divergence minimization.
\newblock In \emph{Advances in Neural Information Processing Systems}, pages
  271--279, 2016.

\bibitem[P{\'o}czos and Schneider(2011)]{poczos2011estimation}
Barnab{\'a}s P{\'o}czos and Jeff Schneider.
\newblock On the estimation of alpha-divergences.
\newblock In \emph{AISTATS}, 2011.

\bibitem[Radford et~al.(2015)Radford, Metz, and
  Chintala]{radford2015unsupervised}
Alec Radford, Luke Metz, and Soumith Chintala.
\newblock Unsupervised representation learning with deep convolutional
  generative adversarial networks.
\newblock \emph{arXiv preprint arXiv:1511.06434}, 2015.

\bibitem[R{\'e}nyi et~al.(1961)]{renyi1961measures}
Alfr{\'e}d R{\'e}nyi et~al.
\newblock On measures of entropy and information.
\newblock In \emph{Proceedings of the Fourth Berkeley Symposium on Mathematical
  Statistics and Probability, Volume 1: Contributions to the Theory of
  Statistics}. The Regents of the University of California, 1961.

\bibitem[Salimans et~al.(2016)Salimans, Goodfellow, Zaremba, Cheung, Radford,
  and Chen]{salimans2016improved}
Tim Salimans, Ian Goodfellow, Wojciech Zaremba, Vicki Cheung, Alec Radford, and
  Xi~Chen.
\newblock Improved techniques for training gans.
\newblock In \emph{NIPS}, 2016.

\bibitem[Sugiyama et~al.(2008)Sugiyama, Nakajima, Kashima, Buenau, and
  Kawanabe]{sugiyama2008direct}
Masashi Sugiyama, Shinichi Nakajima, Hisashi Kashima, Paul~V Buenau, and
  Motoaki Kawanabe.
\newblock Direct importance estimation with model selection and its application
  to covariate shift adaptation.
\newblock In \emph{NIPS}, 2008.

\bibitem[Sutherland et~al.(2016)Sutherland, Tung, Strathmann, De, Ramdas,
  Smola, and Gretton]{sutherland2016generative}
Dougal~J Sutherland, Hsiao-Yu Tung, Heiko Strathmann, Soumyajit De, Aaditya
  Ramdas, Alex Smola, and Arthur Gretton.
\newblock Generative models and model criticism via optimized maximum mean
  discrepancy.
\newblock \emph{arXiv preprint arXiv:1611.04488}, 2016.

\bibitem[Tibshirani(1996)]{tibshirani1996regression}
Robert Tibshirani.
\newblock Regression shrinkage and selection via the lasso.
\newblock \emph{Journal of the Royal Statistical Society. Series B
  (Methodological)}, pages 267--288, 1996.

\bibitem[Warde-Farley and Bengio(2016)]{warde2016improving}
David Warde-Farley and Yoshua Bengio.
\newblock Improving generative adversarial networks with denoising feature
  matching.
\newblock 2016.

\bibitem[Wu et~al.(2016)Wu, Burda, Salakhutdinov, and
  Grosse]{wu2016quantitative}
Yuhuai Wu, Yuri Burda, Ruslan Salakhutdinov, and Roger Grosse.
\newblock On the quantitative analysis of decoder-based generative models.
\newblock \emph{arXiv preprint arXiv:1611.04273}, 2016.

\bibitem[Yamada et~al.(2013{\natexlab{a}})Yamada, Kimura, Naya, and
  Sawada]{yamada2013change}
Makoto Yamada, Akisato Kimura, Futoshi Naya, and Hiroshi Sawada.
\newblock Change-point detection with feature selection in high-dimensional
  time-series data.
\newblock In \emph{IJCAI}, 2013{\natexlab{a}}.

\bibitem[Yamada et~al.(2013{\natexlab{b}})Yamada, Suzuki, Kanamori, Hachiya,
  and Sugiyama]{yamada2013relative}
Makoto Yamada, Taiji Suzuki, Takafumi Kanamori, Hirotaka Hachiya, and Masashi
  Sugiyama.
\newblock Relative density-ratio estimation for robust distribution comparison.
\newblock \emph{Neural computation}, 25\penalty0 (5):\penalty0 1324--1370,
  2013{\natexlab{b}}.

\bibitem[Yamada et~al.(2018)Yamada, Umezu, Fukumizu, and
  Takeuchi]{yamada2018post}
Makoto Yamada, Yuta Umezu, Kenji Fukumizu, and Ichiro Takeuchi.
\newblock Post selection inference with kernels.
\newblock In \emph{AISTATS}, 2018.

\bibitem[Zaremba et~al.(2013)Zaremba, Gretton, and Blaschko]{zaremba2013b}
Wojciech Zaremba, Arthur Gretton, and Matthew Blaschko.
\newblock B-test: {A} non-parametric, low variance kernel two-sample test.
\newblock In \emph{NIPS}, 2013.

\end{thebibliography}
\bibliographystyle{plainnat}

\end{document}